\documentclass[]{fairmeta}
\usepackage{makecell}
\usepackage{wrapfig}
\usepackage{tabularx}
\usepackage{textcomp}
\usepackage{stfloats}
\usepackage{url}
\usepackage{verbatim}
\usepackage{titlesec}
\usepackage{tocloft}
\usepackage{adjustbox}
\usepackage{multirow}
\usepackage{pifont}
\usepackage[sc]{mathpazo}
\usepackage{tikz}
\usepackage{comment}
\usepackage{amsmath,amssymb}
\usepackage{colortbl}
\usepackage{natbib}
\usepackage{color}
\usepackage{booktabs} 
\usepackage{hyperref}
\usepackage{graphicx}
\usepackage{subcaption}
\RequirePackage{xspace}
\makeatletter
\DeclareRobustCommand\onedot{\futurelet\@let@token\@onedot}
\def\@onedot{\ifx\@let@token.\else.\null\fi\xspace}
\usepackage[most]{tcolorbox}
\usepackage{array}
\usepackage{siunitx}
\usepackage[table]{xcolor}
\usepackage{caption}
\definecolor{headerpurple}{HTML}{d8d2fc}
\definecolor{rowgray}{gray}{0.95}
\usepackage{CJKutf8}

\makeatother

\definecolor{adptorange}{RGB}{248, 205, 172}
\definecolor{cmpblue}{RGB}{189, 215, 238}

\definecolor{our_red}{RGB}{232,157,160}
\definecolor{our_blue}{RGB}{136,206,230}
\definecolor{our_orange}{RGB}{246,200,168}
\definecolor{our_green}{RGB}{178,211,164}

\definecolor{attn_code0}{RGB}{247,215,200}
\definecolor{attn_code1}{RGB}{238,169,139}
\definecolor{mlp_code0}{RGB}{204,201,221}
\definecolor{mlp_code1}{RGB}{102,95,153}
\definecolor{mygray}{HTML}{f0f0f0}

\definecolor{token_blue}{RGB}{84, 120, 140}

\usepackage{bbding}
\usepackage{fontawesome}
\usepackage{float}

\newlength\savewidth

\newcolumntype{x}[1]{>{\centering\arraybackslash}p{#1pt}}
\newcolumntype{y}[1]{>{\raggedright\arraybackslash}p{#1pt}}
\newcolumntype{z}[1]{>{\raggedleft\arraybackslash}p{#1pt}}

\renewcommand{\paragraph}[1]{\vspace{1.25mm}\noindent\textbf{#1}}

\usepackage{algorithm}
\usepackage{listings}

\definecolor{codeblue}{rgb}{0.25, 0.5, 0.5}
\definecolor{codekw}{rgb}{0.35, 0.35, 0.75}
\lstdefinestyle{Pytorch}{
    language = Python,
    backgroundcolor = \color{white},
    basicstyle = \fontsize{9pt}{8pt}\selectfont\ttfamily\bfseries,
    columns = fullflexible,
    aboveskip=1pt,
    belowskip=1pt,
    breaklines = true,
    captionpos = b,
    commentstyle = \color{codeblue},
    keywordstyle = \color{codekw},
}

\definecolor{green}{HTML}{009000}
\definecolor{red}{HTML}{ea4335}

\title{AlayaVista: Streaming World Modeling from Panoramic States to Perspective Video}
\author[1,2,*]{Jiaming Tan}
\author[1,2]{Mingliang Zhai}
\author[1,3]{Zhen Li}
\author[2, \dagger]{Yuwei Wu}
\author[1, \dagger, \ddagger]{Chuanhao Li}
\author[1, \dagger]{Kaipeng Zhang}

\affiliation[1]{Alaya Lab}
\affiliation[2]{Beijing Institute of Technology}
\affiliation[3]{The University of Tokyo}

\newcommand{\modelname}{AlayaVista}

\abstract{
Interactive video world models must maintain broad scene context under camera motion while producing high-fidelity observations with low latency.
Existing approaches face a representation trade-off: perspective models operate on local views and must preserve off-screen content over long rollouts, whereas broader spatial coverage is typically obtained by synthesizing full-sphere videos or constructing explicit 3D representations.
Motivated by the complementary roles of global context and selective local acuity in visual perception, we present \modelname{}, a camera-controllable streaming video world model that decouples panoramic world evolution from perspective observation synthesis.
Given a single perspective image, \modelname{} constructs a $360^\circ$ scene prior using a pretrained panorama expansion model and then evolves the scene as a camera-conditioned panoramic latent state.
A latent viewport renderer maps this state to the requested perspective video latents, while a perspective refiner restores details, suppresses artifacts, and performs super-resolution.
To support efficient streaming, we adapt the panoramic generator to chunk-autoregressive generation and distill both panoramic generation and perspective refinement into few-step processes.
To provide the supervision required by this design, we construct MUGEN, a large-scale real-world panoramic video dataset containing 1,318 hours of videos at resolutions of at least 4K, together with rich semantic and geometric annotations.
The system is trained on MUGEN and the panoramic subset of Sekai2.
Experiments validate \modelname{} in visual quality, camera controllability, long-horizon stability, and end-to-end streaming efficiency.
By modeling global dynamics in panoramic latent space and allocating high-fidelity synthesis only to the requested perspective viewport, \modelname{} balances spatial coverage, output quality, and computational efficiency.
}

\github{\url{https://alaya-lab.github.io/\modelname}}
\Code{\url{https://github.com/AlayaLab/\modelname}}
\contact{wuyuwei@bit.edu.cn, chuanhao.li@shanda.com, kaipeng.zhang@shanda.com}
\date{\today}

\begin{document}
\maketitle

% $\ddagger$ {Project lead}, 
\begingroup
\renewcommand{\thefootnote}{}
\footnotetext{* Work done during internship at Alaya Lab, $\dagger$ {Corresponding author}, $\ddagger$ {Project lead}}
\endgroup

\begin{figure*}[!h]
    % \vspace{-2cm}
    \centering
    \includegraphics[width=1\linewidth]{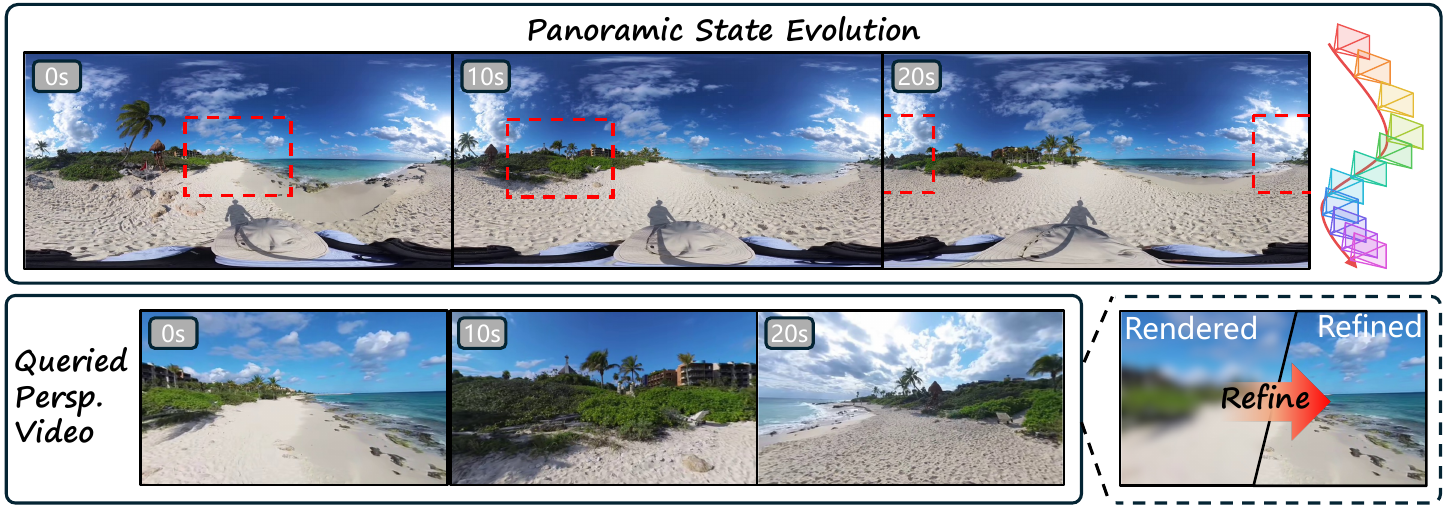}
    \caption{\textbf{From panoramic states to perspective video.}
            \modelname{} evolves camera-conditioned panoramic states, rendering queried viewports (dashed boxes).
            Local refinement enhances visual details (bottom-right).}
    \label{fig:placeholder}
\end{figure*}

\section{Introduction}
\label{sec:introduction}

\noindent
\textit{\textbf{The visual field has boundaries, whereas the visual world has none.}}
\par

\noindent\hfill
\textit{--- James J. Gibson, \emph{The Perception of the Visual World} (1950)}
\par

As Gibson distinguished the bounded visual field from the visual world \cite{gibson1950perception}, an interactive world model should distinguish what the user currently sees from the environment it seeks to represent.
Video world models aim to transform generative video from a passive medium into an interactive environment that responds to user control.
Given an initial visual observation and a camera trajectory, such a model should reveal unseen regions, preserve a coherent scene as the viewpoint changes, and provide continuous visual feedback.
Recent camera-controllable world models have made rapid progress toward long-horizon generation and low-latency streaming \cite{wang2026matrixgame3,wonder2026,chen2026reworld,team2026alayaworldfull,evoke2026}.
A practical system, however, must simultaneously maintain off-screen context, follow extended camera motion, generate high-fidelity observations, and remain computationally efficient.
These requirements create a tension between the spatial scope of the internal world representation and the cost of synthesizing its visible observations.

Most interactive video world models represent world evolution as a sequence of perspective frames because these frames directly match the observations presented to the user.
At any moment, however, a perspective frame captures only a local portion of the surrounding environment.
Once content leaves the current viewport, its appearance and spatial structure must be preserved through temporal context or auxiliary memory so that they can be recovered when the camera returns.
Recent systems therefore employ long-context attention, bounded caches, landmark banks, or explicit spatial memories to support scene recall across extended rollouts \cite{wang2026matrixgame3,chen2026reworld,wang2026mirage,Mao_2026_CVPR, team2026alayaworldv11}.
Although effective, these mechanisms tightly couple the synthesis of the current observation with the storage, retrieval, and updating of previously observed content.

Panoramic representations offer a complementary design by preserving complete angular coverage around the current camera pose.
Recent panoramic video generators and world models exploit this property to improve scene coverage, spherical consistency, camera controllability, and long-horizon exploration \cite{yin2025panoworldx,ji2025campvg,liu2026omniroam,jiang2026panoworld}.
When full-sphere video is treated as the final output, however, high-fidelity generation and refinement must ultimately cover the entire sphere, even though an interactive user observes only one perspective viewport at a time.
Other approaches retain world-space information through Gaussian splats, point clouds, meshes, or spatial memories and render observations from the resulting representation \cite{zhou2026moverse,wang2026mirage,tencent2026hyworld2}.
Such methods can provide strong geometric persistence and revisit consistency, but introduce additional stages for geometric lifting, scene construction, memory maintenance, or rendering.
This raises a central question: can a video world model retain broad visual context without synthesizing every direction at display quality or first constructing an explicit 3D world representation?

Human visual perception suggests a useful computational principle for resolving this trade-off.
Psychophysical studies show that observers can rapidly infer coarse scene layout and global ecological properties from low-spatial-frequency and peripheral information, whereas fine-grained details are acquired selectively through high-acuity central vision \cite{schyns1994blobs,greene2009recognition,larson2009contributions}.
Studies of natural behavior further suggest that detailed visual information is often sampled just in time and in coordination with ongoing actions \cite{hayhoe2003visual}.
These findings do not imply that the brain maintains a pixel-perfect panoramic image of the surrounding environment.
Rather, they motivate an asymmetric global-to-local computation in which a compact representation maintains broad scene context while high-fidelity processing is allocated to the observation currently being queried.

Motivated by this principle, we present \modelname{}, a camera-controllable streaming video world model that decouples panoramic world evolution from perspective observation synthesis.
Given a single perspective image, \modelname{} first uses a pretrained panorama expansion model to construct a complete $360^\circ$ scene prior \cite{tencent2026hyworld2}.
A camera-controllable panoramic video generator then evolves the scene in latent space according to the target camera trajectory.
We refer to the resulting latent sequence as a \emph{panoramic state}: an omnidirectional, camera-centered dynamic representation rather than an explicit metric 3D map.
The panoramic state preserves full angular context at each modeled time step, but is not decoded as the final display-quality output.
Instead, a learned latent viewport renderer maps the panoramic state and the target perspective camera parameters to low-resolution perspective video latents.
A perspective video refiner then restores fine details, suppresses visual artifacts, and performs super-resolution.
By selecting the viewport before high-fidelity synthesis, \modelname{} concentrates expensive computation on the observation presented to the user rather than on the entire sphere.

Realizing this design requires training data that jointly capture panoramic appearance, dynamic real-world content, long temporal context, and controllable camera motion.
Existing generation-oriented panoramic datasets provide captioned videos, but rarely combine large scale, minute-level duration, high resolution, and explicit camera trajectories \cite{wang2024dvd,xia2025panowan}.
Perception-oriented panoramic datasets provide tracking or segmentation annotations, but are not designed to train camera-controllable generative world models \cite{huang2023vot,zhang2025leader360v}.
To provide the supervision required by \modelname{}, we construct MUGEN, a large-scale real-world panoramic video dataset tailored to interactive world modeling.
MUGEN contains 1,318 hours of standardized one-minute panoramic clips at resolutions of at least 4K, covering diverse real-world environments and camera motions.
Each clip is paired with natural-language descriptions and structured semantic attributes, together with geometric annotations including camera trajectories, depth maps, and instance masks.
We further curate MUGEN-HQ, a 300-hour subset selected for visual quality, semantic diversity, and camera-motion diversity.
We train \modelname{} using MUGEN together with the panoramic subset of Sekai2 \cite{he2026sekai2}, combining complementary sources of panoramic video supervision.

To convert the resulting high-quality generator into a streamable world model, we adopt a progressive training strategy.
The panoramic generator first learns long-window, camera-conditioned scene dynamics and is then adapted to chunk-autoregressive rollout.
It is subsequently distilled into a few-step generator to reduce the cost of continuous state evolution.
The latent viewport renderer is trained separately as an interface between panoramic and perspective latent spaces.
After the upstream components are fixed, the perspective video refiner is trained for detail enhancement and super-resolution and is likewise distilled into a few-step model.
This staged procedure separates global dynamics learning, viewport projection, local enhancement, and deployment-time acceleration while enabling efficient end-to-end streaming.

We evaluate \modelname{} in terms of perspective-video quality, camera controllability, long-horizon stability, viewpoint-revisit consistency, and end-to-end streaming efficiency.
The evaluation examines not only the quality of the final perspective observations, but also whether the panoramic state follows the requested trajectory and remains stable during autoregressive rollout.
Experiments validate the effectiveness of the proposed global-state and local-observation decomposition for coherent camera-controlled generation.
They further demonstrate that high-fidelity computation can be concentrated on the requested viewport without requiring final-quality synthesis over the complete sphere.

In summary, our contributions are threefold:
\begin{itemize}
    \item We present \modelname{}, a single-image streaming video world model that represents world evolution through panoramic video latents and decouples global panoramic dynamics from local perspective observation synthesis through latent viewport rendering and perspective refinement.
    \item To support the training of \modelname{}, we construct MUGEN, a large-scale real-world panoramic video dataset containing 1,318 hours of videos at resolutions of at least 4K with rich semantic and geometric annotations, together with the 300-hour high-quality subset MUGEN-HQ.
    \item We develop a progressive training pipeline that combines long-window dynamics learning, chunk-autoregressive rollout, separately trained latent rendering, and few-step distillation for efficient end-to-end streaming under camera control.
\end{itemize}
\section{Related Work}
\label{sec:related_work}

\subsection{Panoramic Video Dataset}
\label{sec:related_panoramic_data}

Existing panoramic video datasets can be broadly grouped into generation-oriented, geometry-oriented, and perception-oriented resources.
WEB360~\cite{wang2024dvd} and PanoVid~\cite{xia2025panowan} provide captioned panoramic clips for text-conditioned video synthesis.
The 360-1M dataset~\cite{wallingford2024image360} mines cross-view correspondences from one million $360^\circ$ videos for large-scale novel-view synthesis and scene imagination.
PanFlow~\cite{zhang2026panflow} emphasizes motion-rich panoramic videos with frame-level camera poses and optical flow for controllable motion generation.
Geometry-oriented resources provide denser spatial supervision: PanoGeo~\cite{jiang2026panoworld} unifies depth, trajectories, and prompts across real and synthetic data; World360~\cite{li2026panoworldreal} combines real panoramic aerial videos with simulated sequences; and Holo360D~\cite{ou2026holo360d} pairs continuous panoramic trajectories with LiDAR-derived geometry.
Perception-oriented panoramic datasets~\cite{huang2023vot,xu2025vots,yan2024panovos,zhang2025leader360v} mainly target tracking, segmentation, and multi-task scene understanding rather than generative world modeling.
Sekai2~\cite{he2026sekai2} contributes long-form real-world videos, camera trajectories, temporally structured annotations, and panoramic sequences containing loops and revisits.
Despite this progress, few datasets jointly provide large-scale real-world panoramic video, minute-level duration, high resolution, natural scene dynamics, continuous camera trajectories, and rich semantic and geometric annotations.
We introduce MUGEN to address this gap, providing 1,318 hours of panoramic videos at resolutions of at least 4K, together with temporally aligned captions, camera trajectories, depth maps, and instance masks.

\subsection{Panoramic Video Generation}
\label{sec:related_panoramic_generation}

Panoramic video generation has progressed from adapting perspective diffusion priors to ERP geometry toward constructing controllable and explorable $360^\circ$ visual worlds.
Panorama-specific diffusion methods~\cite{wang2024dvd,park2025spherediff,xie2025videopanda,xia2025panowan,hirschorn2026spherope} introduce spherical latent representations, multi-view attention, latitude--longitude-aware operations, or sphere-native positional encodings to handle distortion, longitude periodicity, and seam continuity.
Perspective-to-panorama lifting offers another route.
Imagine360~\cite{tan2024imagine360} expands a perspective anchor into an immersive panoramic video, while CubeComposer~\cite{li2026cubecomposer} performs autoregressive generation over cube faces and time to produce native 4K panoramic video.
ViewPoint~\cite{fang2025viewpoint} improves the transfer of perspective video priors to panoramic synthesis, whereas DynamicScaler~\cite{liu2025dynamicscaler} targets scalable high-resolution panoramic generation.

Recent work increasingly treats panoramic generation as a substrate for world exploration.
Image as a World~\cite{gui2025imageworld} integrates single-image world initialization, viewpoint exploration, and temporal continuation within a panoramic video framework.
PanoWorld-X~\cite{yin2025panoworldx} introduces a sphere-aware architecture for explorable panoramic worlds, while CamPVG~\cite{ji2025campvg} designs panoramic camera conditioning for trajectory-controlled generation.
OmniRoam~\cite{liu2026omniroam} combines a fast panoramic preview with temporal extension and spatial refinement for long-horizon wandering.
PanoWorld: Geometry-Consistent Panoramic Video World Modeling~\cite{jiang2026panoworld} regularizes panoramic generation with depth and point trajectories.
Pantheon360~\cite{chen2026pantheon360} couples panoramic diffusion with an explicit 3D cache, while PanoWorld: Real-World Panoramic Generation~\cite{li2026panoworldreal} introduces dense panoramic ray conditioning and geometry-aware memory for long-range exploration.
Most of these methods retain panoramic RGB video as the principal visual output, even when it is later reused for exploration or reconstruction.
In contrast, \modelname{} treats panoramic video latents as internal dynamic world states, maps only the requested viewport into perspective latent space, and postpones display-quality synthesis until after view selection.

\subsection{Perspective World Modeling}
\label{sec:related_perspective_world_modeling}

Most interactive video world models operate directly in perspective space, since their generated frames are also the observations presented to the user.
Camera-controlled perspective video methods~\cite{wang2024motionctrl,he2025cameractrl,xu2024camco,he2025cameractrl2,ren2025gen3c} inject camera trajectories through motion features, ray embeddings, or geometry-aware conditions, and interactive systems extend this capability to sequential autoregressive rollouts.
AlayaWorld~\cite{team2026alayaworldfull} combines chunk-wise generation with bounded temporal context and geometry-aligned spatial memory.
Wonder~\cite{wonder2026} converts a bidirectional video prior into a causal streaming generator, while ReWorld~\cite{chen2026reworld} uses bounded KV caching and a pose-indexed landmark bank for long-horizon recall.
Because each perspective frame covers only a local field of view, these models must preserve off-screen content through context compression, retrieval, or explicit memory~\cite{xiao2025worldmem,wu2025longtermspatialmemory,hong2025relic,xu2026ucm,wang2026mirage}.
To reduce response latency, recent streaming systems~\cite{yin2025causvid,team2026alayaworldfull,wonder2026,chen2026reworld} further combine chunk-autoregressive generation with few-step distillation.
This perspective-space formulation directly matches the final output format, but tightly couples world evolution, memory maintenance, and observation synthesis.

A line of work separates the world representation from the perspective observations synthesized from it.
Explicit 3D world-building methods~\cite{zhang2025worldprompter,li2026pano2world,su2026geniesimpanoworld,fang2026spatialcrafter} construct Gaussian splats, meshes, point clouds, or spatial proxies before view synthesis.
HY-World~2.0~\cite{tencent2026hyworld2} expands a single image into a panorama and builds navigable Gaussian and mesh representations.
MoVerse~\cite{zhou2026moverse} lifts a panorama into a persistent Gaussian scaffold and renders perspective video observations from it.
Generative rendering methods~\cite{liang2025diffusion,huang2026generative,lin2026alayarendererflash,zhang2026renderflow} instead use diffusion or flow models to translate structured geometry, G-buffers, or coarse renderings into photorealistic video.
These approaches provide explicit spatial structure or strong rendering controllability, but require a scene representation or rendering interface.
In contrast, \modelname{} neither evolves the world solely through local perspective frames nor constructs an explicit 3D asset.
It maintains a panoramic video latent as the internal dynamic state, projects only the requested viewport into perspective latent space, and performs high-fidelity refinement after view selection.
% Requires amsmath, amssymb, and graphicx; define \modelname{} in the preamble.
% Editorial clarification: the Haar loss below uses mean-reduced errors,
% following the detailed draft's stated coefficient-count interpretation.
% This reduction convention has not been verified against the training code.
% The supplied drafts do not identify the provenance of the frozen
% quality-refiner checkpoint used to supply pseudo-targets.

\section{\modelname}
\label{sec:method}

\subsection{Overview}
\label{sec:method_overview}

\begin{figure}[t]
    \centering
    \includegraphics[width=\linewidth]{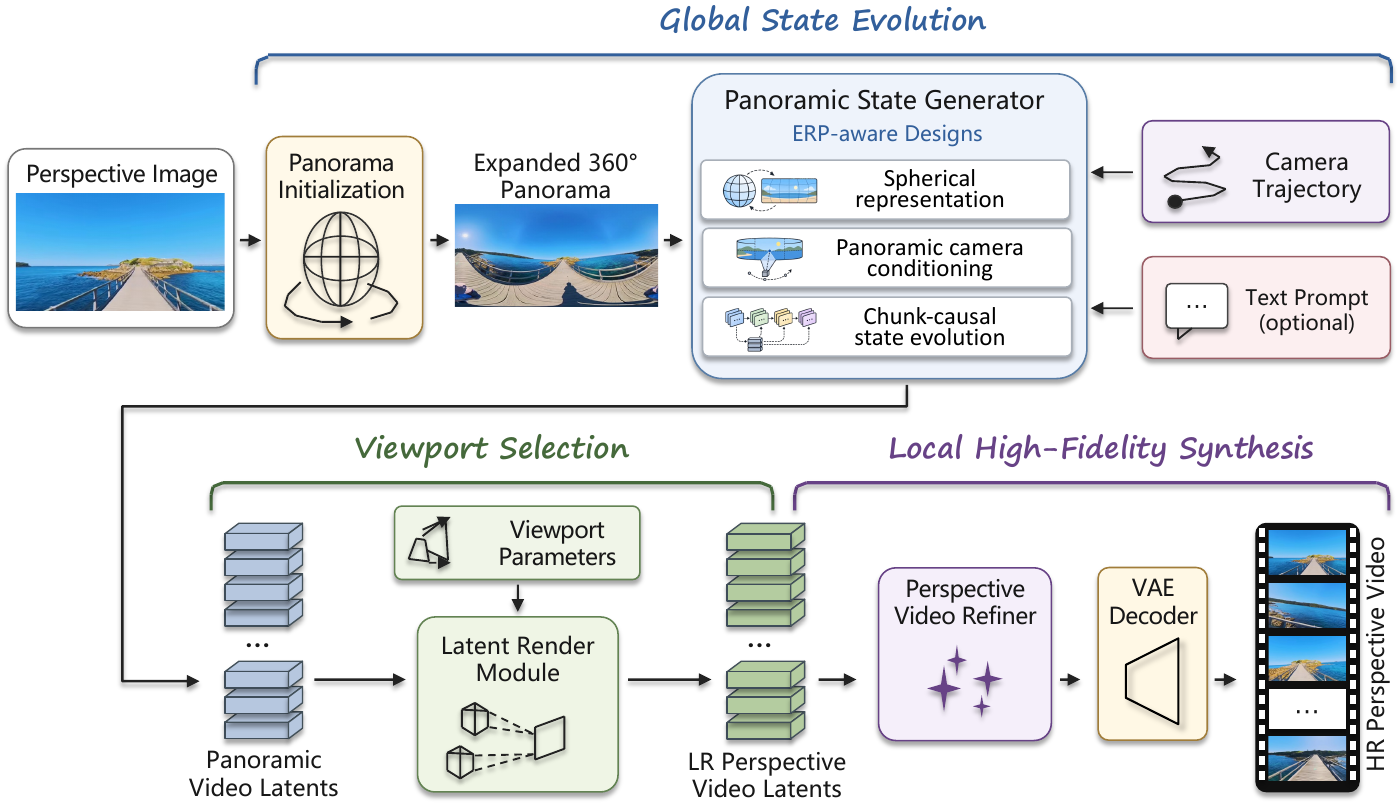}
    \caption{\textbf{Overview of \modelname{}.}
    Camera-conditioned panoramic states are converted into perspective videos through latent rendering, spatial upsampling, and local refinement, followed by RGB decoding.}
    \label{fig:method_pipeline}
\end{figure}

\modelname{} separates omnidirectional world evolution from high-fidelity perspective observation synthesis.
As illustrated in Fig.~\ref{fig:method_pipeline}, the framework consists of four functional modules: a panorama initializer, an ERP-aware panoramic state generator, a geometry-guided latent render module, and a perspective video refiner that integrates latent spatial upsampling with carrier-conditioned generative refinement.
Given a perspective image $\mathbf{I}_0$, a target camera trajectory $\boldsymbol{\Pi}$, per-frame viewport parameters $\boldsymbol{\kappa}$, and an optional text condition $\mathbf{c}$, the inference pipeline is
\begin{equation}
\begin{aligned}
    \mathbf{P}_0
    &= \mathcal{O}(\mathbf{I}_0), \\
    \mathbf{Z}^{\mathrm{pan}}_{1:T}
    &= \mathcal{G}_{\theta}(\mathbf{Z}^{\mathrm{init}},\boldsymbol{\Pi},\mathbf{c}), \\
    \mathbf{Z}^{\mathrm{view,LR}}_{1:T}
    &= \mathcal{R}_{\psi}(\mathbf{Z}^{\mathrm{pan}}_{1:T},\boldsymbol{\kappa}), \\
    \mathbf{C}_{1:T}
    &= \mathcal{U}_{\omega}(\mathbf{Z}^{\mathrm{view,LR}}_{1:T}), \\
    \mathbf{Z}^{\mathrm{view,HR}}_{1:T}
    &= \mathcal{F}_{\phi}(\mathbf{C}_{1:T},\mathbf{c}), \\
    \mathbf{V}^{\mathrm{HR}}
    &= \mathcal{D}_{\mathrm{Wan}}(\mathbf{Z}^{\mathrm{view,HR}}_{1:T}).
\end{aligned}
\label{eq:inference_pipeline}
\end{equation}
Here, $\mathcal{O}$ denotes the pretrained panorama expansion model, $\mathbf{P}_0$ is its ERP output, and $\mathbf{Z}^{\mathrm{init}}$ is a clean initialization block encoded from that panorama.
The generator $\mathcal{G}_{\theta}$ evolves the panoramic state, and the render module $\mathcal{R}_{\psi}$ selects the perspective observation.
Within the perspective video refiner, $\mathcal{U}_{\omega}$ performs deterministic latent upsampling and $\mathcal{F}_{\phi}$ performs carrier-conditioned generative refinement.
The symbols $\theta$, $\psi$, $\omega$, and $\phi$ denote the parameter sets of these networks.
The fixed decoder $\mathcal{D}_{\mathrm{Wan}}$ converts the final perspective latents into RGB.
The symbols $\mathcal{G}_{\theta}$ and $\mathcal{F}_{\phi}$ denote complete sampling procedures with random inputs suppressed; $\mathcal{F}_{\phi}$ includes carrier re-noising, window-wise refinement, and overlap blending when needed.
Their single-evaluation flow-velocity predictors are denoted by $\mathbf{v}_{\theta}$ and $\mathbf{v}_{\phi}$, respectively.

Throughout this section, $t$ indexes latent time and $T$ denotes the number of latent slices, not the number of RGB frames.
The trajectory $\boldsymbol{\Pi}$ provides camera-to-world poses at latent timestamps, whereas $\boldsymbol{\kappa}$ specifies the viewport at every RGB frame.
We suppress temporal subscripts when an expression applies to an entire latent sequence or a refinement window.
RGB resolutions are reported as width $\times$ height, while latent grids are reported as height $\times$ width.

We refer to $\mathbf{Z}^{\mathrm{pan}}_{1:T}$ as the \emph{panoramic state trajectory}, an omnidirectional visual representation centered at the evolving camera pose rather than an explicit metric 3D map.
The upsampled perspective latent $\mathbf{C}$ serves as a \emph{structural carrier}, providing both the re-noised initialization and the clean condition for refinement.
After panorama initialization and encoding, generation, rendering, upsampling, and refinement all operate in latent space.
Only the final refined perspective latent is decoded into RGB during deployment.

\subsection{Panorama Initialization}
\label{sec:panorama_initialization}

We employ the pretrained panorama model from HY-World~2.0~\cite{tencent2026hyworld2} to expand the input perspective image into a complete $2{:}1$ equirectangular projection (ERP) panorama.
This fixed module synthesizes the unobserved surroundings and provides a full-sphere scene prior for subsequent generation.
The expanded panorama specifies the initial scene, while its camera-conditioned temporal evolution is learned by the panoramic state generator.
The initialization block is constructed using the repeated-image encoding procedure described in Sec.~\ref{sec:panoramic_generator_training}.

\subsection{ERP-Aware Panoramic State Generator}
\label{sec:panoramic_state_generator}

The panoramic generator is initialized from \texttt{Wan2.2-TI2V-5B}~\cite{wan2025wan} and operates on the 48-channel latents of \texttt{WanVideoVAE38}.
We retain the pretrained diffusion transformer's text conditioning and flow-matching parameterization, while adapting its positional encoding, VAE boundary handling, and camera-conditioning pathway to panoramic geometry.

\paragraph{Spherical representation.}
Following SpheRoPE~\cite{hirschorn2026spherope}, we replace the native width-axis rotary positional encoding (RoPE) with a two-path spherical construction.
Higher-frequency channels use integer longitudinal harmonics, making their rotary phases periodic across the ERP seam.
Lower-frequency channels use continuous spherical coordinates proportional to $\cos\varphi\cos\lambda$ and $\cos\varphi\sin\lambda$, where $\lambda$ and $\varphi$ denote longitude and latitude.
These coordinates remain seam-continuous and reduce longitude dependence near the poles.
The temporal and height-axis RoPE bands are unchanged.
We additionally apply longitude-circular padding to the panoramic VAE's spatial convolutions and resampling operators during encoding and decoding, while retaining the original temporal and latitude padding.
The VAE weights remain fixed, and spatial tiling is disabled to avoid introducing artificial internal boundaries.

\paragraph{Panoramic camera conditioning.}
We adapt Unified Camera Positional Encoding (UCPE)~\cite{zhang2025ucpe} into a parallel attention pathway specialized for ERP tokens.
Camera poses are expressed relative to the first frame, and moving trajectories are rescaled using their mean displacement between adjacent latent frames; near-static trajectories are left unchanged.
For each ERP token, we construct its spherical viewing ray and a local ray coordinate frame using the corresponding camera pose.
Let $\mathbf{T}_i\in\mathbb{R}^{4\times4}$ be the homogeneous world-to-ray transform of token $i$.
For an attention head of dimension $d_h$, divisible by four, define
\begin{equation}
    \boldsymbol{\Gamma}_i
    =\mathbf{I}_{d_h/4}\otimes\mathbf{T}_i,
    \label{eq:pano_ucpe_transform}
\end{equation}
where $\mathbf{I}_{d_h/4}$ is the identity matrix and $\otimes$ denotes the Kronecker product.
For query, key, and value vectors $\mathbf{q}_i,\mathbf{k}_j,\mathbf{v}_j\in\mathbb{R}^{d_h}$, the camera branch applies
\begin{equation}
\begin{aligned}
    \overline{\mathbf{q}}_i&=\boldsymbol{\Gamma}_i^{\mathsf T}\mathbf{q}_i, \\
    \overline{\mathbf{k}}_j&=\boldsymbol{\Gamma}_j^{-1}\mathbf{k}_j, \\
    \overline{\mathbf{v}}_j&=\boldsymbol{\Gamma}_j^{-1}\mathbf{v}_j.
\end{aligned}
\label{eq:pano_ucpe_qkv}
\end{equation}
The attended feature is mapped into the query ray frame by $\boldsymbol{\Gamma}_i$ and projected back to the backbone width before residual fusion with native self-attention.
The transformed query--key product contains $\boldsymbol{\Gamma}_i\boldsymbol{\Gamma}_j^{-1}$, and therefore depends on the relative ray-frame transform $\mathbf{T}_i\mathbf{T}_j^{-1}$.
The pathway uses compressed attention features and zero-initialized output projections in every transformer block, preserving the pretrained network at initialization.

\paragraph{Chunk-causal state evolution.}
For streaming generation, the model uses bidirectional attention within each latent chunk and causal attention across chunks.
Both native self-attention and panoramic UCPE attention maintain key--value (KV) caches with aligned temporal offsets.
This allows each chunk to reuse generated panoramic features with their camera-relative geometry, without recomputing the entire history.

\subsection{Latent Render Module}
\label{sec:latent_viewport_renderer}

The render module converts a panoramic latent trajectory into the requested perspective observation without first decoding the complete ERP video.
Its viewport condition specifies yaw, pitch, and horizontal field of view within the ERP frame being queried, at every RGB timestamp.
These viewport parameters select observation directions within the panoramic representation, while $\boldsymbol{\Pi}$ controls the camera poses along which that representation evolves.
At the operating resolution, the module maps latents of a $960\times480$ ERP video to those of a $512\times288$ perspective video, changing the latent grid from $30\times60$ to $18\times32$ while preserving the 48 channels and temporal length.

The target mapping is defined by the decode--project--encode reference operator
\begin{equation}
    \mathbf{Z}^{\mathrm{view},*}
    =\mathcal{E}_{\mathrm{Wan}}\!\left(
        \mathcal{W}_{\boldsymbol{\kappa}}\!\left(
            \mathcal{D}_{\mathrm{pan}}(\mathbf{Z}^{\mathrm{pan}})
        \right)
    \right),
    \label{eq:renderer_reference_operator}
\end{equation}
where $\mathcal{W}_{\boldsymbol{\kappa}}$ denotes pixel-space gnomonic projection, $\mathcal{D}_{\mathrm{pan}}$ the ERP-aware decoder, and $\mathcal{E}_{\mathrm{Wan}}$ the standard perspective Wan VAE encoder.
The superscript $*$ marks the low-resolution perspective supervision target, not the final high-resolution output.
At inference, a learned latent-space mapping approximates this RGB-space operator.
Analytic viewport geometry determines sampling locations, while the network learns VAE-specific nonlinear corrections.

The module combines a factorized video transformer with a compact local resampling adapter.
Perspective queries are constructed from geometry-aligned ERP neighbors, fractional sampling offsets, and spherical position and distortion features.
Because each non-initial Wan latent slice represents four RGB frames, we retain four viewport pose anchors per slice instead of collapsing camera motion into a single pose.
The transformer cross-attends to the ERP tokens using a soft geometric bias derived from the projection footprint and camera-motion sweep, followed by perspective spatial and temporal self-attention.
A local adapter additionally learns geometry-conditioned corrections from nearby $4\times4$ ERP latent neighborhoods.
The output combines a bilinear reference sample with global and local learned residuals:
\begin{equation}
\begin{aligned}
    \mathbf{Z}^{\mathrm{view,LR}}
    ={}&\mathcal{S}_{\boldsymbol{\kappa}}(\mathbf{Z}^{\mathrm{pan}}) \\
    &+\Delta\mathbf{Z}^{\mathrm{global}}
    +\Delta\mathbf{Z}^{\mathrm{local}}.
\end{aligned}
\label{eq:renderer_residual_output}
\end{equation}
Here, $\mathcal{S}_{\boldsymbol{\kappa}}$ denotes geometry-conditioned bilinear latent sampling at a designated viewport anchor.
The residuals $\Delta\mathbf{Z}^{\mathrm{global}}$ and $\Delta\mathbf{Z}^{\mathrm{local}}$ are predicted by the global transformer and local adapter, respectively, and share the shape of the perspective output latent.
The module therefore retains an explicit geometric sampling prior without requiring the full panoramic RGB decode--project--encode path at deployment.

\subsection{Perspective Video Refiner}
\label{sec:perspective_video_refiner}

The perspective video refiner converts low-resolution viewport latents into high-resolution observations through two internal components: a deterministic latent spatial upsampler and a carrier-conditioned generative refinement network.
The upsampler establishes the target spatial grid and structural carrier, while the generative network synthesizes appearance details before the final RGB decoding.

\paragraph{Latent spatial upsampling.}
\label{sec:latent_spatial_upsampler}
The deterministic upsampler $\mathcal{U}_{\omega}$ prepares a high-resolution structural carrier from the low-resolution render output.
Using an LTX-2-inspired residual and spatial PixelShuffle organization, we train the upsampler specifically in Wan latent space rather than directly applying an LTX upsampler checkpoint.
Three-dimensional residual blocks aggregate local spatiotemporal information, while spatial PixelShuffle doubles the spatial resolution without changing the temporal length.

The upsampler combines a fixed nearest-neighbor carrier with a learned residual:
\begin{equation}
\begin{aligned}
    \Delta\mathbf{Z}^{\mathrm{raw}}
    &=\mathcal{A}_{\omega}\!\left(
        \boldsymbol{\mu}+\boldsymbol{s}\odot\mathbf{Z}^{\mathrm{view,LR}}
    \right), \\
    \mathbf{C}
    &=\operatorname{NN}_{2\times}(\mathbf{Z}^{\mathrm{view,LR}})
      +\Delta\mathbf{Z}^{\mathrm{raw}}\oslash\boldsymbol{s}.
\end{aligned}
\label{eq:upsampler_nearest_residual}
\end{equation}
Here, $\boldsymbol{\mu},\boldsymbol{s}\in\mathbb{R}^{48}$ are the fixed per-channel mean and standard deviation used by the Wan VAE, broadcast over time and spatial positions.
The symbols $\odot$ and $\oslash$ denote elementwise multiplication and division, and $\operatorname{NN}_{2\times}$ denotes twofold spatial nearest-neighbor upsampling.
The network $\mathcal{A}_{\omega}$ is the learned branch of $\mathcal{U}_{\omega}$ and predicts the high-resolution residual $\Delta\mathbf{Z}^{\mathrm{raw}}$ in the unnormalized VAE latent domain.
Dividing this residual by $\boldsymbol{s}$ converts it back to normalized latent units before addition to the fixed carrier.
The final projection of $\mathcal{A}_{\omega}$ is initialized to zero, so training starts from the nearest-neighbor mapping.
The output grid increases from $18\times32$ to $36\times64$, corresponding to a resolution change from $512\times288$ to $1024\times576$.
This output is a high-resolution structural carrier rather than the final detailed video: it carries layout, motion, and coarse appearance, while ambiguous high-frequency textures are delegated to the generative refiner.

\paragraph{Carrier-conditioned causal refinement.}
The refiner is initialized from \texttt{Wan2.2-TI2V-5B}~\cite{wan2025wan} and processes the upsampled carrier in chunks of four latent slices.
For chunk $n$, sampling starts by re-noising its carrier:
\begin{equation}
    \mathbf{X}_{n,\sigma_0}
    =(1-\sigma_0)\mathbf{C}_n+\sigma_0\boldsymbol{\epsilon}_n,
    \label{eq:refiner_initialization}
\end{equation}
where $\boldsymbol{\epsilon}_n$ is Gaussian noise.
The clean carrier is appended after the noisy tokens as an aligned reference, using the same spatiotemporal positions and a zero diffusion timestep.
For $n>1$, the preceding refined chunk is prepended as a clean prefix, giving
\begin{equation}
    [\widehat{\mathbf{Z}}_{n-1},\mathbf{X}_{n,\sigma},\mathbf{C}_n],
    \label{eq:refiner_chunk_input}
\end{equation}
with the prefix omitted for the first chunk.
Only the current target tokens are decoded; attention is bidirectional within a chunk and causal across chunks.

\paragraph{Four-step causal inference.}
After distilled, each chunk is denoised with four step.
The result $\widehat{\mathbf{Z}}_n$ is emitted and reused as the clean prefix for chunk $n+1$.
Long videos are obtained by directly concatenating the refined chunks.

\subsection{Training Stages}
\label{sec:training_stages}

\begin{figure}[t]
    \centering
    \includegraphics[width=\linewidth]{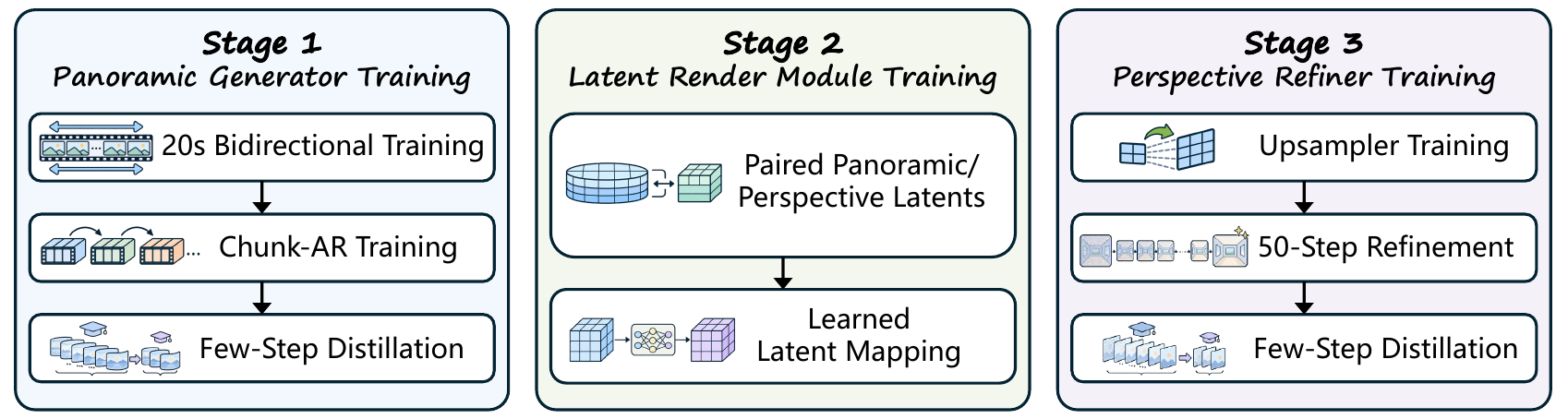}
    \caption{\textbf{Training stages of \modelname{}.}
    Panoramic generation, latent rendering, and perspective refinement are trained separately; perspective refinement begins with latent upsampler training.}
    \label{fig:training_stage}
\end{figure}

As illustrated in Fig.~\ref{fig:training_stage}, training is organized into three groups: panoramic generator training, render module training, and perspective video refiner training.
The third group comprises latent spatial upsampler training, multi-step quality training, and few-step distillation, in that order.
The panorama initializer and both VAE backbones remain fixed throughout training.
The panoramic generator and render module are frozen before training the perspective branch, and the upsampler is frozen after its training substep.
The complete pipeline is not jointly optimized end to end.

\subsubsection{Panoramic Generator Training}
\label{sec:panoramic_generator_training}

The panoramic generator is trained through bidirectional adaptation, chunk-autoregressive training, and few-step distillation.

\paragraph{Bidirectional panoramic adaptation.}
We first adapt the pretrained video model to panoramic generation using non-causal temporal attention and a short-to-long training schedule.
A stabilized 10-second checkpoint initializes the causal generator, while a separate bidirectional copy is extended to 20-second clips and retained as the long-window score model for subsequent distribution matching.
The model is trained with image-, video-, and text-conditioned examples.
Let $\mathbf{Z}$ be a clean panoramic latent video, $\boldsymbol{\epsilon}$ a tensor of independent standard Gaussian noise, and $\mathbf{M}$ a binary mask shaped like $\mathbf{Z}$, with ones on prediction targets and zeros on clean conditioning states.
For a sampled noise level $\sigma\in[0,1]$, the masked flow path is
\begin{equation}
    \mathbf{Z}_{\sigma}
    =(1-\sigma\mathbf{M})\odot\mathbf{Z}
      +\sigma\mathbf{M}\odot\boldsymbol{\epsilon}.
    \label{eq:pano_flow_path}
\end{equation}
Thus, target states are interpolated toward noise, while conditioning states remain clean.
We optimize
\begin{equation}
    \mathcal{L}_{\mathrm{FM}}
    =\mathbb{E}\!\left[
        \frac{w_{\mathrm{Wan}}(\sigma)}{\|\mathbf{M}\|_1}
        \left\|\mathbf{M}\odot\left(
        \mathbf{v}_{\theta}(\mathbf{Z}_{\sigma},\sigma,\boldsymbol{\Pi},\mathbf{c})
        -(\boldsymbol{\epsilon}-\mathbf{Z})\right)\right\|_{F}^{2}
    \right],
    \label{eq:pano_flow_loss}
\end{equation}
where $w_{\mathrm{Wan}}(\sigma)$ is the Wan timestep weight, $\|\cdot\|_F^2$ sums squared tensor entries, and $\|\mathbf{M}\|_1$ counts the supervised entries.
The expectation is over training examples, conditioning configurations, sampled noise levels, and Gaussian noise.
Every training example contains at least one supervised target state.

\paragraph{Chunk-autoregressive training.}
We convert the generator to block-causal attention with four latent slices per chunk, corresponding to approximately 16 RGB frames at 16 FPS, except for the VAE's initial-frame convention.
At single-image inference, the panorama is repeated for 13 RGB frames and encoded into a clean four-slice initialization block.
Let $\mathbf{B}^{\mathrm{pan}}_n$ denote the $n$-th generated chunk and $N$ the number of generated chunks, excluding the initialization block.
The complete panoramic sequence consists of the initialization block followed by the generated chunks.
Generation factorizes as
\begin{equation}
\begin{aligned}
    &p_{\theta}(\mathbf{B}^{\mathrm{pan}}_{1:N}
        \mid\mathbf{P}_0,\boldsymbol{\Pi},\mathbf{c}) \\
    &\qquad=\prod_{n=1}^{N}
    p_{\theta}(\mathbf{B}^{\mathrm{pan}}_n
        \mid\mathcal{M}_n,\mathbf{P}_0,
        \boldsymbol{\Pi}_{[\leq n]},\mathbf{c}).
\end{aligned}
\label{eq:chunk_autoregressive_factorization}
\end{equation}
Here, $\mathcal{M}_n$ is the cached history containing the initialization block and preceding generated chunks in the native and UCPE attention pathways.
The notation $\boldsymbol{\Pi}_{[\leq n]}$ includes all camera poses up to the end of chunk $n$, rather than the first $n$ individual poses.
The current chunk is denoised jointly without access to future chunks.
Training begins with teacher-forced histories augmented by latent corruption and replayed prediction errors, and subsequently introduces detached self-resampled and teacher-generated rollout histories.
These history constructions expose the model to imperfect deployment-like context, while the initialization block remains protected and no gradient is propagated through the constructed history.

\paragraph{Few-step distillation.}
We first initialize a four-step student through consistency distillation on transitions from a 50-level causal teacher, using an exponential-moving-average (EMA) student as the lower-noise target.
We then apply on-policy Self-Forcing++~\cite{cui2025self} and distribution matching~\cite{yin2024dmd2} to trajectories generated by the student's own chunk-wise rollout.
The frozen real-score model and trainable fake-score model are initialized from the 20-second bidirectional checkpoint.
A score window is selected from each student rollout, with its leading context block detached from the student gradient.
For a student window $\widehat{\mathbf{Z}}^{\mathrm{stu}}$, we form a perturbed sample at score noise level $\tau\in(0,1)$:
\begin{equation}
    \widetilde{\mathbf{Z}}_{\tau}
    =(1-\tau)\operatorname{sg}
      (\widehat{\mathbf{Z}}^{\mathrm{stu}})
      +\tau\boldsymbol{\epsilon}',
    \label{eq:pano_score_perturbation}
\end{equation}
where $\boldsymbol{\epsilon}'$ is independent Gaussian noise and $\operatorname{sg}$ denotes stop-gradient.
Let $\widehat{\mathbf{Z}}^{\mathrm{real}}$ and $\widehat{\mathbf{Z}}^{\mathrm{fake}}$ be the clean-latent predictions obtained from the real and fake score models using this same perturbed window, noise level, and camera/text conditions.
Each is obtained by subtracting $\tau$ times the corresponding velocity prediction from $\widetilde{\mathbf{Z}}_{\tau}$.
For each sample, the normalized update direction and its surrogate objective are
\begin{equation}
\begin{aligned}
    \mathbf{g}
    &=\frac{\widehat{\mathbf{Z}}^{\mathrm{fake}}
             -\widehat{\mathbf{Z}}^{\mathrm{real}}}
        {\operatorname{mean}\!\left|
            \widehat{\mathbf{Z}}^{\mathrm{stu}}
            -\widehat{\mathbf{Z}}^{\mathrm{real}}
        \right|+\varepsilon}, \\
    \mathcal{L}_{\mathrm{DMD}}
    &=\frac{1}{2}\left\|
        \widehat{\mathbf{Z}}^{\mathrm{stu}}
        -\operatorname{sg}\!\left(
            \widehat{\mathbf{Z}}^{\mathrm{stu}}-\mathbf{g}
        \right)
    \right\|_F^2.
\end{aligned}
\label{eq:pano_dmd_loss}
\end{equation}
Here, $\operatorname{mean}$ averages latent entries per sample, and $\varepsilon>0$ is a scalar numerical stabilizer, distinct from Gaussian noise tensors.
The fake score is fitted to detached student-generated samples using flow matching.
A sparse spectral anchor reuses early cached features to constrain low-frequency layout and color during later rollout steps, leaving the complementary high-frequency attention features unchanged.
The final student retains the panoramic geometry modules, camera pathway, and chunk-causal KV caches.

\subsubsection{Render Module Training}
\label{sec:latent_viewport_renderer_training}

We train the render module with cached panoramic--perspective latent pairs generated by the frozen reference operator in Eq.~\eqref{eq:renderer_reference_operator}.
The inputs are derived from real panoramic videos, and the sampled viewport trajectories include static, smooth, rapid, full-spin, and seam-crossing motion.
Supervision combines channel-normalized latent reconstruction, first- and second-order temporal differences, and decoded-video losses for perceptual similarity, image gradients, and high-frequency temporal consistency.
We first optimize the global transformer and then freeze it while fitting the zero-initialized local resampling adapter.
The target is the deterministic viewport transformation, not unconstrained appearance generation.
The RGB decode--project--encode reference is used for supervision, while deployment retains only latent-space geometric sampling and learned correction.

\subsubsection{Perspective Video Refiner Training}
\label{sec:perspective_refiner_training}

This training group prepares the high-resolution perspective branch after the panoramic generator and render module have been fixed.
It consists of three substeps: latent spatial upsampler training, multi-step quality training, and few-step distillation.
Only the upsampler is optimized in the first substep; it is then frozen while the generative refinement network is trained and distilled.

\paragraph{Latent spatial upsampler training.}
\label{sec:latent_spatial_upsampler_training}
We first train $\mathcal{U}_{\omega}$ to reproduce a deterministic high-resolution carrier rather than regress directly to ambiguous native high-resolution textures.
For each low-resolution perspective latent, the carrier target is
\begin{equation}
    \mathbf{Z}^{\mathrm{car}}
    =\mathcal{E}_{\mathrm{Wan}}\!\left(
        \operatorname{Bicubic}_{2\times}\!\left[
            \mathcal{D}_{\mathrm{Wan}}(\mathbf{Z}^{\mathrm{view,LR}})
        \right]
    \right),
    \label{eq:upsampler_carrier_target}
\end{equation}
where $\operatorname{Bicubic}_{2\times}$ applies twofold spatial bicubic resizing independently to each decoded frame without resampling time.
The upsampler prediction $\mathbf{C}=\mathcal{U}_{\omega}(\mathbf{Z}^{\mathrm{view,LR}})$ is supervised by $\mathbf{Z}^{\mathrm{car}}$; the former is the learned output and the latter is the fixed reference target.
Inputs are mixed from real-video ERP latents, bidirectional panoramic generations, and four-step panoramic generations, all processed by the fixed render module.
Each target is constructed from its own low-resolution input, maintaining a deterministic training correspondence for every source.
The objective combines normalized latent reconstruction, multiscale structure preservation, temporal and high-frequency consistency, and auxiliary decoded-video perceptual supervision.
The VAE decode--resize--encode operation is used to construct supervision and is replaced by the learned upsampler at deployment.
After this substep, $\mathcal{U}_{\omega}$ is frozen for both subsequent substeps.

\paragraph{Multi-step quality adaptation.}
We first train a normal-step carrier-conditioned refiner to recover high-frequency appearance while preserving the layout and motion specified by the upsampled carrier.
Training mixes carriers rendered from real videos, bidirectional panoramic generations, and four-step panoramic generations.
Real examples use VAE-encoded high-resolution perspective videos as targets, while generated panoramic inputs are paired with fixed high-quality pseudo-targets.
Let $\mathbf{Y}_n$ and $\widehat{\mathbf{Z}}_n$ denote the target and predicted clean latent chunks, respectively.
The quality objective is
\begin{equation}
\begin{aligned}
    \mathcal{L}_{\mathrm{Q}}
    ={}&\mathcal{L}_{\mathrm{FM}}
      +\lambda_{\mathrm{str}}\mathcal{L}_{\mathrm{str}}
      +\lambda_{\mathrm{hf}}\mathcal{L}_{\mathrm{hf}}
      +\lambda_{\mathrm{perc}}\mathcal{L}_{\mathrm{perc}} \\
     &+\lambda_{\mathrm{temp}}\mathcal{L}_{\mathrm{temp}}
      +\lambda_{\mathrm{bnd}}\mathcal{L}_{\mathrm{bnd}},
\end{aligned}
\label{eq:refiner_quality_objective}
\end{equation}
where $\mathcal{L}_{\mathrm{FM}}$ is the Wan flow-matching objective,
$\mathcal{L}_{\mathrm{str}}=\|\mathcal{H}_{\mathrm{LL}}(\widehat{\mathbf{Z}}_n)-\mathcal{H}_{\mathrm{LL}}(\mathbf{C}_n)\|_1$ anchors low-frequency structure to the carrier, and
$\mathcal{L}_{\mathrm{hf}}=\|\mathcal{H}_{\mathrm{HF}}(\widehat{\mathbf{Z}}_n)-\mathcal{H}_{\mathrm{HF}}(\mathbf{Y}_n)\|_1$ restores target details.
The decoded perceptual loss $\mathcal{L}_{\mathrm{perc}}$ improves visual quality, while $\mathcal{L}_{\mathrm{temp}}$ matches first- and second-order temporal changes and $\mathcal{L}_{\mathrm{bnd}}$ constrains the transition between adjacent chunks.

\paragraph{Four-step Self-Forcing distillation.}
The quality model is then used as the frozen real-score model to distill a four-step causal student with Self-Forcing~\cite{huang2025selfforcing}.
The student follows the same clean-prefix and aligned-carrier layout as inference: it rolls out chunk by chunk and reuses each detached prediction as the prefix of the next chunk.
For a perturbed student sample, let $\mathbf{Y}^{\mathrm{real}}_n$ and $\mathbf{Y}^{\mathrm{fake}}_n$ be the clean predictions of the frozen real-score model and the learned fake-score model.
We use the normalized distribution-matching direction
\begin{equation}
\begin{aligned}
    \mathbf{g}_n
    &=\frac{\mathbf{Y}^{\mathrm{fake}}_n-\mathbf{Y}^{\mathrm{real}}_n}
    {\operatorname{mean}|\widehat{\mathbf{Z}}_n-\mathbf{Y}^{\mathrm{real}}_n|+\varepsilon}, \\
    \mathcal{L}_{\mathrm{DMD}}^{\mathrm{ref}}
    &=\frac{1}{2}\left\|\widehat{\mathbf{Z}}_n-
    \operatorname{sg}(\widehat{\mathbf{Z}}_n-\mathbf{g}_n)\right\|_F^2,
\end{aligned}
\label{eq:refiner_dmd_objective}
\end{equation}
and optimize
\begin{equation}
    \mathcal{L}_{\mathrm{SF}}
    =\lambda_{\mathrm{DMD}}\mathcal{L}_{\mathrm{DMD}}^{\mathrm{ref}}
    +\lambda_{\mathrm{str}}\mathcal{L}_{\mathrm{str}}
    +\lambda_{\mathrm{bnd}}\mathcal{L}_{\mathrm{bnd}}
    +\lambda_{\mathrm{rel}}\mathcal{L}_{\mathrm{rel}}.
\label{eq:refiner_self_forcing_objective}
\end{equation}
Here, $\mathcal{L}_{\mathrm{rel}}$ preserves low-frequency motion and high-frequency energy across consecutive chunks.
Training on the student's own generated history aligns the training and inference distributions and reduces long-horizon error accumulation.

\section{Training Data}
\label{sec:training_data}

Training \modelname{} requires panoramic videos that jointly provide realistic scene appearance, long-duration temporal evolution, diverse camera motion, and temporally aligned semantic and geometric supervision.
To meet these requirements, we introduce MUGEN, a large-scale real-world panoramic video dataset designed for camera-controllable world modeling.
We further incorporate the panoramic subset of Sekai2 \cite{he2026sekai2} as an additional source of panoramic video supervision.
Although MUGEN and Sekai2 are collected from different sources, their panoramic videos are processed using the same ERP-based hierarchical annotation pipeline.
This unified design produces consistent training records containing panoramic videos, temporally aligned captions, and per-frame camera information.

\subsection{MUGEN Collection and Curation}
\label{sec:mugen_collection}

\begin{figure}
    \centering
    \includegraphics[width=\linewidth]{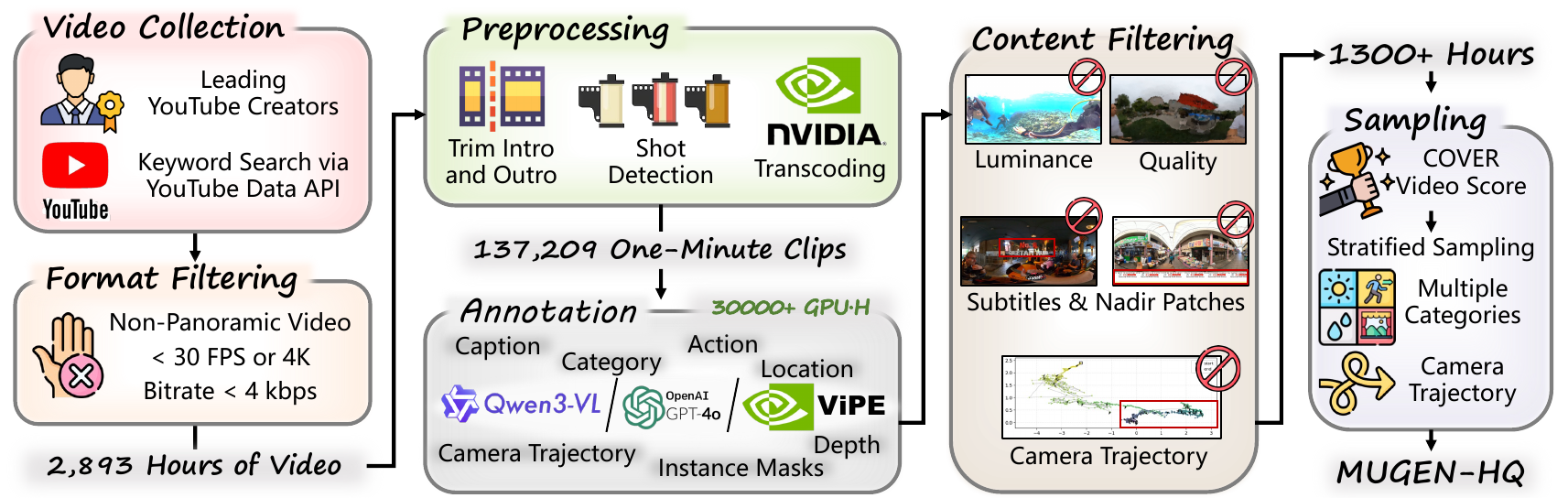}
    \caption{\textbf{MUGEN curation pipeline.} We collect, preprocess, annotate, and filter panoramic videos to construct MUGEN, then obtain MUGEN-HQ through quality-aware stratified sampling.}
    \label{fig:data_pipeline}
\end{figure}

\paragraph{Video acquisition.}
We use YouTube as the primary source of diverse, long-form panoramic videos captured in real-world environments.
We manually curate representative panoramic-video creators and query the YouTube Data API using keywords such as ``360 tour.''
The retrieved results are manually reviewed to exclude non-real-world content, including simulated, animated, and rendered scenes.
This process yields 9,544 candidate source videos.

We subsequently apply metadata- and format-based filtering to retain high-fidelity panoramic footage.
We remove non-panoramic videos, videos with resolutions below 4K, frame rates below 30 FPS, or insufficient bitrates.
For each retained source, we download the highest available video resolution and audio quality.
After format filtering, the source collection contains 2,893 hours of raw panoramic footage.

\paragraph{Temporal preprocessing.}
We first verify file integrity and remove videos that cannot be decoded reliably.
The first and last minute of each source video are trimmed to remove common introductions, credits, and outros.
Because Internet videos frequently contain edited transitions, we use TransNetV2 \cite{soucek2020transnetv2} to detect shot boundaries and retain only temporally continuous segments.
Each continuous segment is then partitioned into consecutive, non-overlapping 60-second clips, providing a standardized unit for scalable annotation and model training.
The resulting clips are transcoded to H.265 at their original resolution and frame rate, while audio is encoded as AAC at 48 kHz.
This preprocessing stage produces 137,209 one-minute clips totaling approximately 2,287 hours before content filtering.

\paragraph{Quality filtering.}
Raw Internet videos may contain exposure failures, blur, compression artifacts, visible overlays, panoramic stitching artifacts, or unreliable camera motion.
We therefore apply complementary filters covering photometric quality, perceptual quality, visual overlays, and trajectory validity.

We first compute the mean luma value of each clip and discard clips outside the range $[40,215]$ to remove severely underexposed or overexposed content.
We then evaluate perceptual video quality using COVER \cite{he2024cover} and remove clips with scores below 0.7.
To identify subtitles, watermarks, interface elements, and visible nadir patches, we project each ERP clip into six perspective views and inspect them using a vision-language model.
These perspective projections are used only for quality filtering and are not used for semantic caption annotation.
Finally, we discard clips whose estimated camera trajectories contain chaotic paths, temporal discontinuities, or abnormal rotations.

\paragraph{Geometric annotation.}
For camera-control supervision, we use ViPE \cite{huang2025vipe} to estimate per-frame camera trajectories.
ViPE additionally provides depth maps and instance masks for each panoramic clip.
The recovered camera trajectories are used both as continuous control signals and as cues for trajectory-validity filtering.
After all filtering stages, MUGEN contains 1,318 hours of high-quality one-minute panoramic clips collected from 6,446 unique source videos.

\paragraph{MUGEN-HQ.}
For efficient model development and controlled evaluation, we further construct MUGEN-HQ, a 300-hour high-quality subset of MUGEN.
We first rank the retained clips according to their COVER scores and preserve the top 70\% as a quality-filtered candidate pool.
We then perform stratified sampling over scene categories, camera-motion patterns, action types, and weather conditions.
This procedure preserves high visual quality while maintaining diversity in both scene content and controllable camera motion.

\subsection{Unified Panoramic Video Annotation}
\label{sec:panoramic_annotation}

A single clip-level caption is insufficient for streaming world modeling because scene content, subject motion, environmental dynamics, and camera behavior may change substantially over time.
We therefore adopt the hierarchical training-oriented annotation scheme of AlayaWorld \cite{team2026alayaworldfull} and apply it uniformly to the panoramic videos from both MUGEN and Sekai2.
The shared pipeline uses the same panoramic input representation, temporal sampling strategy, annotation fields, and output format for the two datasets.

\paragraph{ERP-based visual input.}
All panoramic videos are represented as unfolded $2{:}1$ equirectangular projection (ERP) sequences.
For semantic annotation, the complete ERP frames are directly provided to the vision-language model and are not decomposed into separate cube faces.
We sample each video at 1--2 FPS and attach an explicit \texttt{[mm:ss]} timestamp to every sampled frame.
The timestamped ERP sequence preserves the full panoramic observation while encouraging the annotation model to identify when scene content, motion, and camera behavior change.

\begin{figure*}[t]
    \centering
    \includegraphics[width=\linewidth]{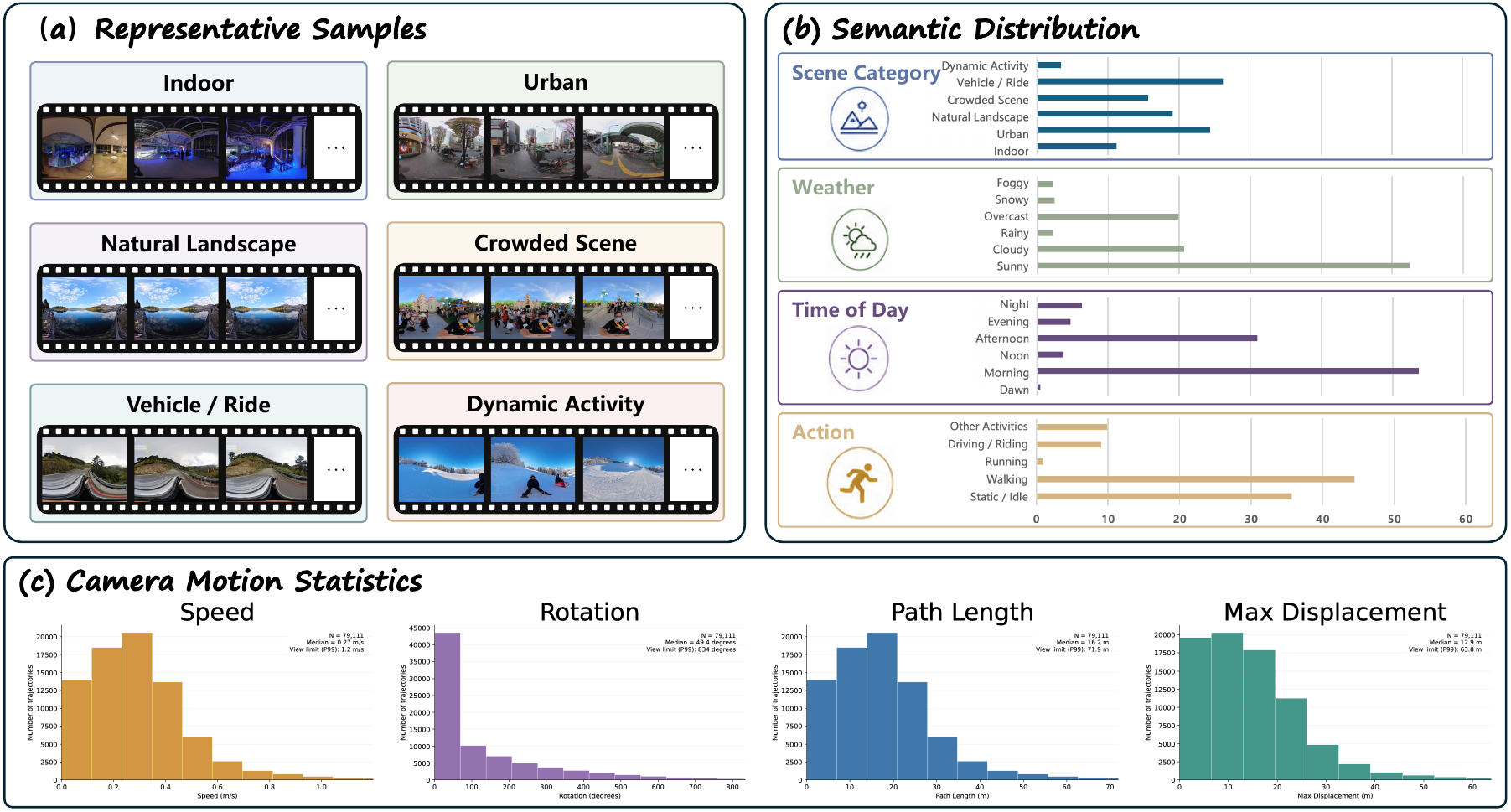}
    \caption{\textbf{MUGEN statistics.} (a) Representative samples, (b) semantic distributions, and (c) camera-motion statistics.}
    \label{fig:data_statistic}
\end{figure*}

\paragraph{Video-level context.}
At the video level, we annotate a compact set of global attributes that remain relatively stable throughout the clip.
These attributes include weather, time of day, location type, camera perspective, camera motion, and video style.
The controlled vocabulary provides global context for text-conditioned generation and also supports dataset analysis and balanced sampling.

\paragraph{Segment-level dynamics.}
Each video is further partitioned into temporally grounded segments represented by \texttt{time\_range\_s}.
For each segment, we generate four factorized semantic tracks:
\texttt{subject\_motion} describes the motion and actions of the primary subjects;
\texttt{environment\_motion} describes changes in surrounding entities, lighting, weather, and other environmental elements;
\texttt{static\_scene} records persistent scene appearance, spatial layout, and visible objects;
and \texttt{camera\_description} describes viewpoint, framing, camera motion, and camera stability.

Explicitly separating subject dynamics from camera behavior is important for camera-controllable generation.
For example, the annotation can distinguish a moving subject from a translating camera, an object rotation from a camera orbit, and a static environment observed under a panning viewpoint.

\paragraph{Prompt construction and camera labels.}
The factorized segment descriptions are fused into a detailed \texttt{full\_prompt} containing 5--9 present-tense sentences.
We additionally generate a concise \texttt{short\_prompt} of 15--45 words for caption dropout and multi-caption augmentation.
Each segment is assigned a \texttt{camera\_path} label from a 16-way camera-motion vocabulary defined independently of subject motion.
The discrete \texttt{camera\_path} label complements the continuous per-frame camera trajectory and provides an interpretable description of the dominant camera behavior.

Applying the same annotation pipeline to MUGEN and Sekai2 avoids source-dependent differences in caption granularity, temporal segmentation, and camera-motion terminology.
Consequently, clips from both datasets can be mixed directly during training without requiring separate text-conditioning formats.

\subsection{Unified Training Corpus}
\label{sec:data_mixture}

The final training corpus combines MUGEN with the panoramic subset of Sekai2 \cite{he2026sekai2}.
MUGEN provides large-scale, high-resolution panoramic videos covering diverse real-world scenes and camera motions.
Sekai2 contributes a complementary source of panoramic sequences and further broadens the scene and trajectory distributions observed during training.

We normalize the two datasets into a common training record containing an ERP video, hierarchical semantic annotations, and per-frame camera information.
Because both datasets use the same ERP-based annotation schema, their video-level attributes, segment-level tracks, prompts, and camera-motion labels share a consistent definition.
\section{Experiments}
\label{sec:experiments}

We evaluate \modelname{} on camera-controlled perspective video synthesis from a single image.
Our experiments comprise quantitative comparisons with two external baselines, followed by qualitative visualizations of camera control and panoramic--perspective correspondence.

\subsection{Quantitative Evaluation}
\label{sec:quantitative_evaluation}

\paragraph{Evaluation protocol.}
We select 200 evaluation cases from MUGEN-HQ.
For each case, we project the original ERP sequence into perspective views to construct a reference video and use its first frame as the input image.
The target perspective camera trajectory is obtained by combining the source camera poses with the corresponding viewport orientations.
All methods receive the same input image and target camera trajectory, with camera coordinates and field-of-view settings converted to their respective interfaces.
Future reference frames and the complete reference panorama are not provided as visual conditions.
Generated and reference videos are evaluated at matched timestamps, using 81 frames per video at a common resolution of 1024 × 576 pixels.

\paragraph{Compared methods.}
We compare against MoVerse~\cite{zhou2026moverse} and HY-World~2.0~\cite{tencent2026hyworld2}, evaluating their final perspective RGB outputs.
For MoVerse, we use the final synthesized perspective video rather than intermediate Gaussian renderings.
For HY-World~2.0, we render its final 3D Gaussian Splatting (3DGS) scene along the target camera trajectory to obtain the perspective video, and do not evaluate intermediate outputs from its world-construction pipeline.
For \modelname{}, we evaluate the final RGB video after latent rendering, spatial upsampling, and perspective refinement.
All methods are therefore compared in the same perspective observation space, irrespective of their internal world representations.

\paragraph{Evaluation metrics.}
For reconstruction-oriented quality, we report FVD~\cite{unterthiner2018towards}, SSIM~\cite{wang2004image}, LPIPS~\cite{zhang2018unreasonable}, and PSNR against reference perspective videos.
For perceptual and temporal quality, we use VBench++~\cite{huang2024vbenchpp} and report Consistency, Quality, and Dynamic scores.
For camera control, following CameraCtrl~\cite{he2025cameractrl}, we estimate the camera trajectory of each output video using ViPE~\cite{huang2025vipe} and compute the translation error (TransErr) against the input trajectory.
All metrics are computed on the final perspective RGB videos.

\paragraph{Quantitative comparison.}
Table~\ref{tab:quantitative_results} summarizes the results on the 200 MUGEN-HQ evaluation cases.
These metrics characterize complementary aspects of the final perspective videos, separating reference agreement from perceptual quality, temporal consistency, apparent motion, and trajectory following.
AlayaVista achieves the best SSIM (0.4616), LPIPS (0.5321), and PSNR (14.108), indicating closer agreement with the reference videos than both baselines. It also obtains the highest Consistency score (0.9240) and a modest improvement in Quality (0.5579), demonstrating favorable temporal consistency and perceptual quality. For camera control, AlayaVista achieves the lowest rotation error (2.132), indicating more accurate tracking of the target camera orientations. The lower Dynamic score (0.9200) may reflect attenuation of local motion during latent rendering and video refinement, suggesting a possible trade-off between motion activity and temporal consistency. Although our translation error (0.03312) is lower than that of HY-World 2.0 (0.03885), it remains higher than MoVerse’s (0.02746), whose explicit 3D Gaussian scaffold may provide stronger geometric guidance during video synthesis.

\begin{table*}[t]
    \centering
    \caption{\textbf{Quantitative comparison on 200 MUGEN-HQ cases.} All metrics are computed on final perspective videos.}
    \label{tab:quantitative_results}
    \small
    \setlength{\tabcolsep}{5pt}
    \resizebox{\textwidth}{!}{%
    \begin{tabular}{lcccccccc}
        \toprule
        Method & SSIM $\uparrow$ & LPIPS $\downarrow$ & Consistency $\uparrow$ & Quality $\uparrow$ & Dynamic $\uparrow$ & PSNR $\uparrow$ & TransErr $\downarrow$ & RotErr $\downarrow$ \\
        \midrule
        MoVerse~\cite{zhou2026moverse}
        & 0.4232 & 0.5583 & 0.8836 & 0.5517 & \textbf{1.0000} & 13.17 & \textbf{0.02746} & $2.887$ \\
        HY-World~2.0~\cite{tencent2026hyworld2}
        & 0.4524 & 0.5417 & 0.8662 & 0.5293 & 0.9850 & 13.54 & 0.03885 & $3.169$ \\
        \textbf{\modelname{} (Ours)}
        & \textbf{0.4616} & \textbf{0.5321} & \textbf{0.9240} & \textbf{0.5579} & 0.9200 & \textbf{14.10} & 0.03312 & $\mathbf{2.132}$ \\
        \bottomrule
    \end{tabular}%
    }
\end{table*}

\subsection{Qualitative Evaluation}
\label{sec:qualitative_evaluation}

We present two complementary visualizations of \modelname{}: camera-controlled perspective video generation and the correspondence between panoramic states and perspective observations.

\begin{figure*}[t]
    \centering
    \includegraphics[width=\linewidth]{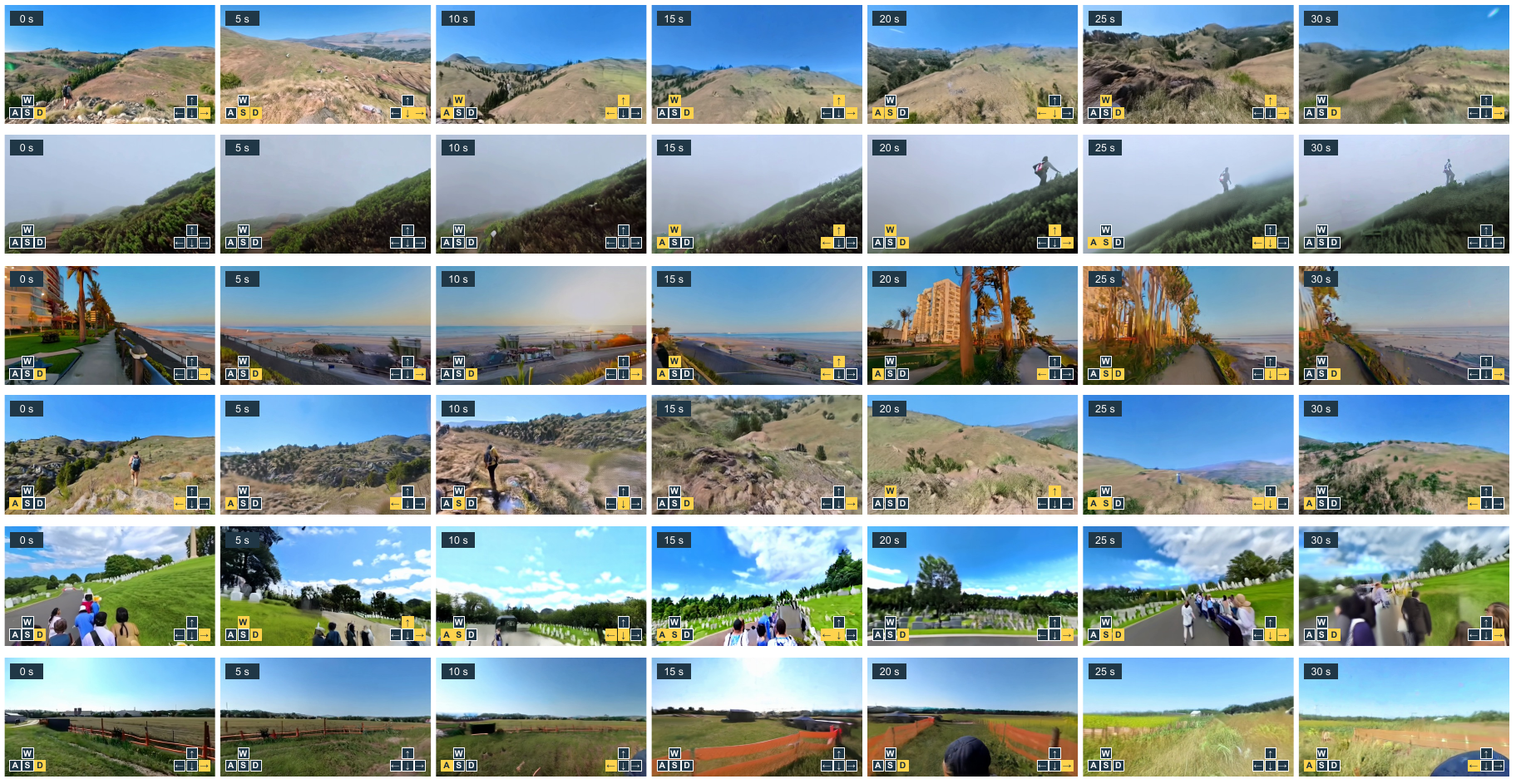}
    \caption{\textbf{Camera-controlled perspective video generation.}
    Input images, target camera trajectories, and corresponding generated frames.}
    \label{fig:perspective_camera_control}
\end{figure*}

\paragraph{Camera-controlled perspective generation.}
Figure~\ref{fig:perspective_camera_control} presents input images, target camera trajectories, and temporally ordered frames from the final perspective videos.
The visualization examines how the displayed observations respond to the requested camera motion.
Changes in framing, relative object positions, and newly revealed regions allow trajectory following to be inspected alongside visual appearance and temporal continuity.
Showing the controls together with the output frames connects camera conditioning to the final user-visible observations and complements the quantitative TransErr evaluation.

\begin{figure*}[t]
    \centering
    \includegraphics[width=\linewidth]{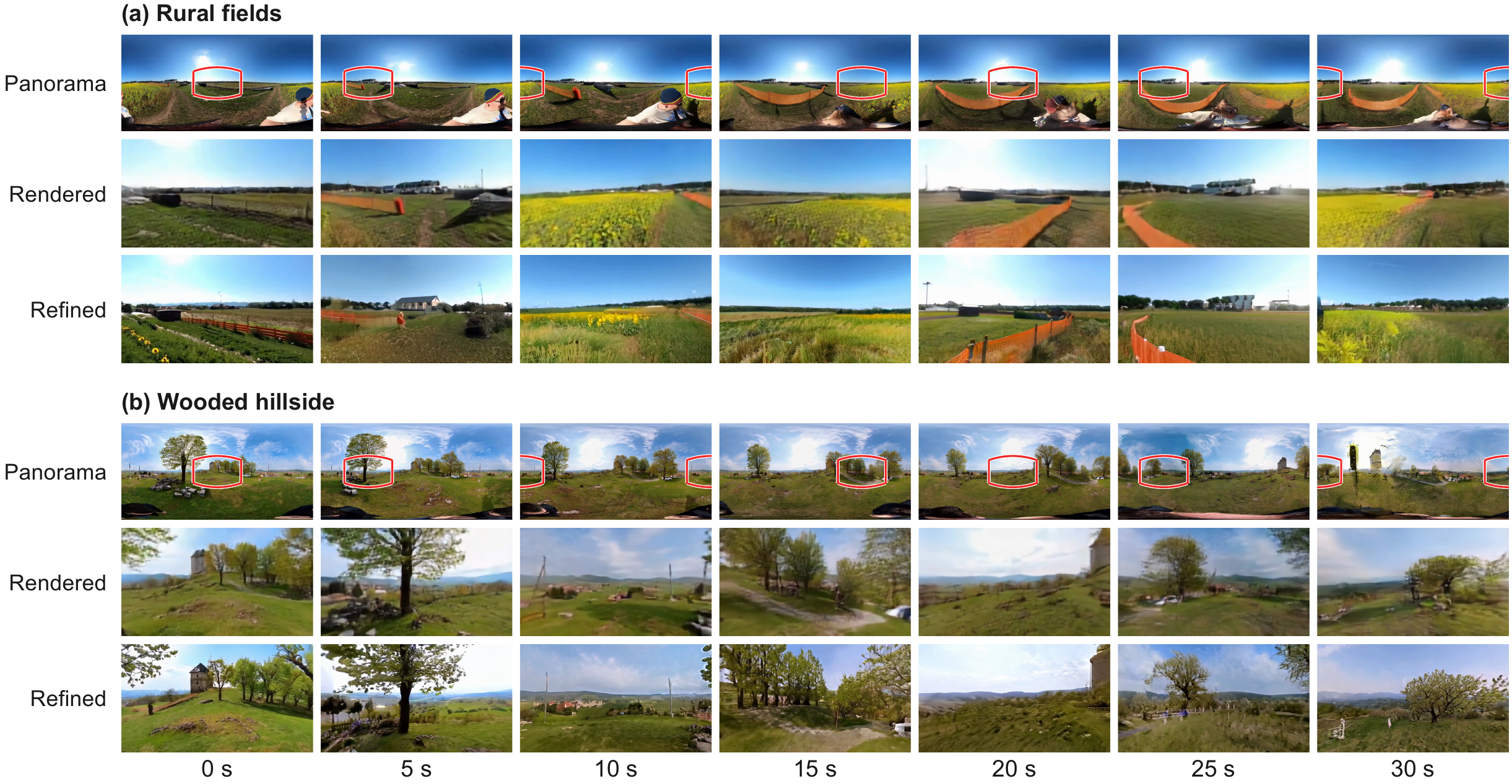}
    \caption{\textbf{Panoramic--perspective correspondence.}
    Panoramic states and refined perspective outputs at matching timestamps, with queried viewports marked.}
    \label{fig:panoramic_perspective_correspondence}
\end{figure*}

\paragraph{Panoramic--perspective correspondence.}
Figure~\ref{fig:panoramic_perspective_correspondence} pairs panoramic state visualizations with the corresponding perspective outputs at matching timestamps.
The queried viewport regions are marked on the ERP frames to indicate which parts of the panoramic state contribute to each local observation.
The panoramic frames are decoded for visualization only and are not intermediate RGB inputs in the deployed pipeline.
Perspective outputs are obtained through latent rendering, spatial upsampling, and generative refinement rather than direct cropping of the displayed ERP frames.
The paired visualization allows scene layout and view coverage to be inspected across the two representations, while also showing the appearance details introduced during perspective synthesis.
Together, the two figures illustrate how \modelname{} connects camera-controlled panoramic state evolution with local perspective observation synthesis.

\section{Conclusion}
\label{sec:conclusion}

In this work, we presented \modelname{}, a camera-controllable streaming video world model that decouples global panoramic world evolution from local perspective observation synthesis.
Starting from a single perspective image, \modelname{} constructs a $360^\circ$ scene prior, evolves it as a camera-conditioned panoramic latent state, and converts the requested viewport into high-fidelity perspective video through latent viewport rendering and perspective refinement.
By treating panoramic video as an internal dynamic state rather than the final output, the proposed global-to-local architecture preserves omnidirectional scene context while concentrating expensive high-fidelity computation on the view presented to the user.
We further enable continuous generation through chunk-autoregressive rollout and few-step distillation of both panoramic generation and perspective refinement.
To support this framework, we introduced MUGEN, a large-scale real-world panoramic video dataset containing 1,318 hours of videos at resolutions of at least 4K with rich semantic and geometric annotations, together with the 300-hour MUGEN-HQ subset.
The complete system is trained using MUGEN and the panoramic subset of Sekai2.
Experiments validate \modelname{} in perspective-video quality, camera controllability, long-horizon stability, viewpoint-revisit consistency, and end-to-end streaming efficiency.
Taken together, these results position panoramic latent states as a practical middle ground between perspective-only world modeling and explicit 3D scene construction, retaining broad visual context without synthesizing the complete sphere at display quality.
Nevertheless, a camera-centered panoramic state does not by itself guarantee persistent spatial memory under large camera translations or strict geometric consistency in highly dynamic scenes, and errors introduced during panorama expansion may propagate through subsequent generation stages.
Future work will explore persistent spatial memory, stronger geometry-aware objectives, and more tightly coupled end-to-end training to support increasingly robust and open-ended interactive world modeling.

\bibliographystyle{abbrv}
\bibliography{references}

@String{CVPR    = {Proceedings of the IEEE/CVF Conference on Computer Vision and Pattern Recognition (CVPR)}}

@String{ICCV    = {Proceedings of the IEEE/CVF International Conference on Computer Vision (ICCV)}}

@String{NeurIPS = {Advances in Neural Information Processing Systems (NeurIPS)}}

@inproceedings{wang2024motionctrl,
  title={Motionctrl: A unified and flexible motion controller for video generation},
  author={Wang, Zhouxia and Yuan, Ziyang and Wang, Xintao and Li, Yaowei and Chen, Tianshui and Xia, Menghan and Luo, Ping and Shan, Ying},
  booktitle={ACM SIGGRAPH 2024 Conference Papers},
  pages={1--11},
  year={2024}
}

@article{he2025cameractrl,
  title={Cameractrl: Enabling camera control for text-to-video generation},
  author={He, Hao and Xu, Yinghao and Guo, Yuwei and Wetzstein, Gordon and Dai, Bo and Li, Hongsheng and Yang, Ceyuan},
  journal={arXiv preprint arXiv:2404.02101},
  year={2024}
}

@article{xu2024camco,
  title={Camco: Camera-controllable 3d-consistent image-to-video generation},
  author={Xu, Dejia and Nie, Weili and Liu, Chao and Liu, Sifei and Kautz, Jan and Wang, Zhangyang and Vahdat, Arash},
  journal={arXiv preprint arXiv:2406.02509},
  year={2024}
}

@inproceedings{he2025cameractrl2,
  title={Cameractrl II: Dynamic scene exploration via camera-controlled video diffusion models},
  author={He, Hao and Yang, Ceyuan and Lin, Shanchuan and Xu, Yinghao and Wei, Meng and Gui, Liangke and Zhao, Qi and Wetzstein, Gordon and Jiang, Lu and Li, Hongsheng},
  booktitle=ICCV,
  pages={13416--13426},
  year={2025}
}

@inproceedings{zhang2025ucpe,
  title={Unified camera positional encoding for controlled video generation},
  author={Zhang, Cheng and Li, Boying and Wei, Meng and Cao, Yan-Pei and Gambardella, Camilo and Phung, Dinh and Cai, Jianfei},
  booktitle=CVPR,
  pages={38027--38037},
  year={2026}
}

@inproceedings{ren2025gen3c,
  title={Gen3c: 3d-informed world-consistent video generation with precise camera control},
  author={Ren, Xuanchi and Shen, Tianchang and Huang, Jiahui and Ling, Huan and Lu, Yifan and Nimier-David, Merlin and M{\"u}ller, Thomas and Keller, Alexander and Fidler, Sanja and Gao, Jun},
  booktitle=CVPR,
  pages={6121--6132},
  year={2025}
}

@article{xiao2025worldmem,
  title={Worldmem: Long-term consistent world simulation with memory},
  author={Xiao, Zeqi and Lan, Yushi and Zhou, Yifan and Ouyang, Wenqi and Yang, Shuai and Zeng, Yanhong and Pan, Xingang},
  journal=NeurIPS,
  volume={38},
  pages={49632--49652},
  year={2026}
}

@article{wu2025longtermspatialmemory,
  title={Video world models with long-term spatial memory},
  author={Wu, Tong and Yang, Shuai and Po, Ryan and Xu, Yinghao and Liu, Ziwei and Lin, Dahua and Wetzstein, Gordon},
  journal=NeurIPS,
  volume={38},
  pages={49371--49393},
  year={2026}
}

@article{huang2025selfforcing,
  title={Self forcing: Bridging the train-test gap in autoregressive video diffusion},
  author={Huang, Xun and Li, Zhengqi and He, Guande and Zhou, Mingyuan and Shechtman, Eli},
  journal=NeurIPS,
  volume={38},
  pages={167283--167308},
  year={2026}
}

@article{yin2024dmd2,
  title={Improved distribution matching distillation for fast image synthesis},
  author={Yin, Tianwei and Gharbi, Micha{\"e}l and Park, Taesung and Zhang, Richard and Shechtman, Eli and Durand, Fredo and Freeman, William T},
  journal=NeurIPS,
  volume={37},
  pages={47455--47487},
  year={2024}
}

@inproceedings{yin2025causvid,
  title={From slow bidirectional to fast causal video generators},
  author={Yin, Tianwei and Zhang, Qiang and Zhang, Richard and Freeman, William T. and Durand, Fr{\'e}do and Shechtman, Eli and Huang, Xun},
  booktitle=CVPR,
  year={2025}
}

@inproceedings{Mao_2026_CVPR,
  title={Yume1.5: A Text-Controlled Interactive World Generation Model},
  author={Mao, Xiaofeng and Li, Zhen and Li, Chuanhao and Xu, Xiaojie and Ying, Kaining and Zhang, Kaipeng},
  booktitle=CVPR,
  pages={7752--7761},
  year={2026}
}

@article{wan2025wan,
  title={Wan: Open and advanced large-scale video generative models},
  author={Wan, Team and Wang, Ang and Ai, Baole and Wen, Bin and Mao, Chaojie and Xie, Chen-Wei and Chen, Di and Yu, Feiwu and Zhao, Haiming and Yang, Jianxiao and others},
  journal={arXiv preprint arXiv:2503.20314},
  year={2025}
}

@article{huang2026generative,
  title={Generative World Renderer},
  author={Huang, Zheng-Hui and Wang, Zhixiang and Tan, Jiaming and Yu, Ruihan and Zhang, Yidan and Zheng, Bo and Liu, Yu-Lun and Chuang, Yung-Yu and Zhang, Kaipeng},
  journal={arXiv preprint arXiv:2604.02329},
  year={2026}
}

@inproceedings{liang2025diffusion,
  title     = {{DiffusionRenderer}: Neural Inverse and Forward Rendering with Video Diffusion Models},
  author    = {Liang, Ruofan and Gojcic, Zan and Ling, Huan and Munkberg, Jacob and Hasselgren, Jon and Lin, Chih-Hao and Gao, Jun and Keller, Alexander and Vijaykumar, Nandita and Fidler, Sanja and Wang, Zian},
  booktitle = CVPR,
  pages     = {26069--26080},
  year      = {2025}
}

@article{zhang2026renderflow,
  title={RenderFlow: Single-Step Neural Rendering via Flow Matching},
  author={Zhang, Shenghao and Liu, Runtao and Schroers, Christopher and Zhang, Yang},
  journal={arXiv preprint arXiv:2601.06928},
  year={2026}
}

@article{cui2025self,
  title={Self-forcing++: Towards minute-scale high-quality video generation},
  author={Cui, Justin and Wu, Jie and Li, Ming and Yang, Tao and Li, Xiaojie and Wang, Rui and Bai, Andrew and Ban, Yuanhao and Hsieh, Cho-Jui},
  journal={arXiv preprint arXiv:2510.02283},
  year={2025}
}

@article{soucek2020transnetv2,
  title   = {{TransNet V2}: An Effective Deep Network Architecture for Fast Shot Transition Detection},
  author  = {Sou{\v{c}}ek, Tom{\'a}{\v{s}} and Loko{\v{c}}, Jakub},
  journal = {arXiv preprint arXiv:2008.04838},
  year    = {2020}
}

@article{huang2025vipe,
  title={{ViPE}: Video Pose Engine for 3D Geometric Perception},
  author={Huang, Jiahui and Zhou, Qunjie and Rabeti, Hesam and Korovko, Aleksandr and Ling, Huan and Ren, Xuanchi and Shen, Tianchang and Gao, Jun and Slepichev, Dmitry and Lin, Chen-Hsuan and Ren, Jiawei and Xie, Kevin and Biswas, Joydeep and Leal-Taixe, Laura and Fidler, Sanja},
  journal={arXiv preprint arXiv:2508.10934},
  year={2025}
}

@inproceedings{he2024cover,
  title={{COVER}: A Comprehensive Video Quality Evaluator},
  author={He, Chenlong and Zheng, Qi and Zhu, Ruoxi and Zeng, Xiaoyang and Fan, Yibo and Tu, Zhengzhong},
  booktitle={Proceedings of the IEEE/CVF Conference on Computer Vision and Pattern Recognition (CVPR) Workshops},
  pages={5799--5809},
  year={2024}
}

@book{gibson1950perception,
  title     = {The Perception of the Visual World},
  author    = {Gibson, James J.},
  year      = {1950},
  publisher = {Houghton Mifflin},
  address   = {Boston}
}

@article{wang2026matrixgame3,
  title   = {{Matrix-Game 3.0}: Real-Time and Streaming Interactive World Model with Long-Horizon Memory},
  author  = {Wang, Zile and Liu, Zexiang and Li, Jaixing and Huang, Kaichen and Xu, Baixin and Kang, Fei and An, Mengyin and Wang, Peiyu and Jiang, Biao and Wei, Yichen and Xietian, Yidan and Pei, Jiangbo and Hu, Liang and Jiang, Boyi and Xue, Hua and Wang, Zidong and Sun, Haofeng and Li, Wei and Ouyang, Wanli and He, Xianglong and Liu, Yang and Li, Yangguang and Zhou, Yahui},
  journal = {arXiv preprint arXiv:2604.08995},
  year    = {2026}
}

@article{wonder2026,
  title   = {Wonder: Video World Model Done Better},
  author  = {Xu, Jiacong and Jiang, Hanwen and Shu, Zhixin and Sunkavalli, Kalyan and Patel, Vishal M. and Mei, Yiqun},
  journal = {arXiv preprint arXiv:2607.26037},
  year    = {2026}
}

@article{chen2026reworld,
  title   = {{ReWorld}: An Interactive World Model with Long-Horizon Memory},
  author  = {Chen, Zhifei and Wang, Luozhou and Shen, Guibao and Yan, Dongyu and Yang, Shuai and Xu, Tianshuo and Du, Yihua and Wang, Wei and Gui, Tianyi and Huang, Lianghua and Chen, Yingcong},
  journal = {arXiv preprint arXiv:2608.23565},
  year    = {2026}
}

@article{wang2026mirage,
  title   = {Latent Spatial Memory for Video World Models},
  author  = {Wang, Weijie and Zhao, Haoyu and Yang, Yifan and Chen, Feng and Zhang, Zeyu and He, Yefei and Duan, Zicheng and Chen, Donny Y. and Yang, Yuqing and Zhuang, Bohan},
  journal = {arXiv preprint arXiv:2606.09828},
  year    = {2026}
}

@article{yin2025panoworldx,
  title   = {{PanoWorld-X}: Generating Explorable Panoramic Worlds via Sphere-Aware Video Diffusion},
  author  = {Yin, Yuyang and Guo, Haoxiang and Liu, Fangfu and Wang, Mengyu and Liang, Hanwen and Li, Eric and Wang, Yikai and Jin, Xiaojie and Zhao, Yao and Wei, Yunchao},
  journal = {arXiv preprint arXiv:2509.24997},
  year    = {2025}
}

@article{ji2025campvg,
  title   = {{CamPVG}: Camera-Controlled Panoramic Video Generation with Epipolar-Aware Diffusion},
  author  = {Ji, Chenhao and Yu, Chaohui and Gao, Junyao and Wang, Fan and Zhao, Cairong},
  journal = {arXiv preprint arXiv:2509.19979},
  year    = {2025}
}

@article{liu2026omniroam,
  title   = {{OmniRoam}: World Wandering via Long-Horizon Panoramic Video Generation},
  author  = {Liu, Yuheng and Lin, Xin and Li, Xinke and Yang, Baihan and Wang, Chen and Sunkavalli, Kalyan and Hold-Geoffroy, Yannick and Tan, Hao and Zhang, Kai and Xie, Xiaohui and Shi, Zifan and Hu, Yiwei},
  journal = {arXiv preprint arXiv:2603.30045},
  year    = {2026}
}

@article{jiang2026panoworld,
  title   = {{PanoWorld}: Geometry-Consistent Panoramic Video World Modeling},
  author  = {Jiang, Le and Bai, Xiangyu and Galoaa, Bishoy and Moezzi, Shayda and Lee, Caleb James and Imtiaz, Tooba and Yeh, Edmund and Dy, Jennifer and Wang, Yanzhi and Ostadabbas, Sarah},
  journal = {arXiv preprint arXiv:2605.15391},
  year    = {2026}
}

@article{zhou2026moverse,
  title   = {{MoVerse}: Real-Time Video World Modeling with Panoramic Gaussian Scaffold},
  author  = {Zhou, Yang and Wang, Ziheng and Lu, Yuqin and Liu, Haofeng and Liang, Jun and He, Shengfeng and Li, Jing},
  journal = {arXiv preprint arXiv:2606.13376},
  year    = {2026}
}

@article{tencent2026hyworld2,
  title   = {{HY-World 2.0}: A Multi-Modal World Model for Reconstructing, Generating, and Simulating 3D Worlds},
  author  = {{Team HY-World} and Cao, Chenjie and Zuo, Xuhui and Wang, Zhenwei and Zhang, Yisu and Wu, Junta and Liu, Zhenyang and Gong, Yuning and Liu, Yang and Yuan, Bo and Zhang, Chao and Li, Coopers and Guo, Dongyuan and Yang, Fan and Zhang, Haiyu and Cao, Hang and Zhu, Jianchen and Lin, Jiaxin and Xiao, Jie and Zhang, Jihong and Yu, Junlin and Wang, Lei and Wang, Lifu and Wang, Lilin and Linus and Chen, Minghui and He, Peng and Zhao, Penghao and Chen, Qi and Chen, Rui and Shao, Rui and Liu, Sicong and Qin, Wangchen and Niu, Xiaochuan and Yuan, Xiang and Sun, Yi and Tang, Yifei and Sun, Yifu and Lian, Yihang and Tan, Yonghao and Liu, Yuhong and Yin, Yuyang and Min, Zhiyuan and Wang, Tengfei and Guo, Chunchao},
  journal = {arXiv preprint arXiv:2604.14268},
  year    = {2026}
}

@inproceedings{wang2024dvd,
  title     = {{360DVD}: Controllable Panorama Video Generation with 360-Degree Video Diffusion Model},
  author    = {Wang, Qian and Li, Weiqi and Mou, Chong and Cheng, Xinhua and Zhang, Jian},
  booktitle = {Proceedings of the IEEE/CVF Conference on Computer Vision and Pattern Recognition},
  pages     = {6913--6923},
  year      = {2024}
}

@article{xia2025panowan,
  title   = {{PanoWan}: Lifting Diffusion Video Generation Models to 360$^\circ$ with Latitude/Longitude-Aware Mechanisms},
  author  = {Xia, Yifei and Weng, Shuchen and Yang, Siqi and Liu, Jingqi and Zhu, Chengxuan and Teng, Minggui and Jia, Zijian and Jiang, Han and Shi, Boxin},
  journal = {arXiv preprint arXiv:2505.22016},
  year    = {2025}
}

@inproceedings{huang2023vot,
  title     = {{360VOT}: A New Benchmark Dataset for Omnidirectional Visual Object Tracking},
  author    = {Huang, Huajian and Xu, Yinzhe and Chen, Yingshu and Yeung, Sai-Kit},
  booktitle = {Proceedings of the IEEE/CVF International Conference on Computer Vision},
  pages     = {20566--20576},
  year      = {2023}
}

@article{zhang2025leader360v,
  title   = {{Leader360V}: The Large-Scale, Real-World 360 Video Dataset for Multi-Task Learning in Diverse Environment},
  author  = {Zhang, Weiming and Xiao, Dingwen and Dai, Aobotao and Liu, Yexin and Pan, Tianbo and Wen, Shiqi and Chen, Lei and Wang, Lin},
  journal = {arXiv preprint arXiv:2506.14271},
  year    = {2025}
}

@article{schyns1994blobs,
  title   = {From Blobs to Boundary Edges: Evidence for Time- and Spatial-Scale-Dependent Scene Recognition},
  author  = {Schyns, Philippe G. and Oliva, Aude},
  journal = {Psychological Science},
  volume  = {5},
  number  = {4},
  pages   = {195--200},
  year    = {1994},
  doi     = {10.1111/j.1467-9280.1994.tb00500.x}
}

@article{greene2009recognition,
  title   = {Recognition of Natural Scenes from Global Properties: Seeing the Forest Without Representing the Trees},
  author  = {Greene, Michelle R. and Oliva, Aude},
  journal = {Cognitive Psychology},
  volume  = {58},
  number  = {2},
  pages   = {137--176},
  year    = {2009},
  doi     = {10.1016/j.cogpsych.2008.06.001}
}

@article{larson2009contributions,
  title   = {The Contributions of Central Versus Peripheral Vision to Scene Gist Recognition},
  author  = {Larson, Adam M. and Loschky, Lester C.},
  journal = {Journal of Vision},
  volume  = {9},
  number  = {10},
  pages   = {1--16},
  year    = {2009},
  doi     = {10.1167/9.10.6}
}

@article{hayhoe2003visual,
  title   = {Visual Memory and Motor Planning in a Natural Task},
  author  = {Hayhoe, Mary M. and Shrivastava, Anurag and Mruczek, Ryan and Pelz, Jeff B.},
  journal = {Journal of Vision},
  volume  = {3},
  number  = {1},
  pages   = {49--63},
  year    = {2003},
  doi     = {10.1167/3.1.6}
}

@article{he2026sekai2,
  title   = {{Sekai2}: From World Exploration to Interactive World Modeling},
  author  = {He, Kang and Peng, Wenshuo and Gao, Zihui and Tan, Jiaming and Zhang, Kaipeng and Ge, Yongtao},
  journal = {arXiv preprint arXiv:2608.09449},
  year    = {2026}
}

@article{team2026alayaworldfull,
  title   = {{AlayaWorld}: Interactive Long-Horizon World Modeling -- Full Technical Report},
  author  = {{AlayaWorld Team} and Zhang, Kaipeng and Li, Chuanhao and Zhan, Yifan and Ge, Yongtao and Yin, Yuanyang and Tan, Jiaming and He, Kang and Fan, Liaoyuan and Zhai, Mingliang and Liu, Ruicong and Xu, Xiaojie and Chu, Xuangeng and Li, Zhen and Lin, Zhengyuan and Wang, Zhixiang and Meng, Zian and Gao, Zihui},
  journal = {arXiv preprint arXiv:2607.18367},
  year    = {2026}
}

@article{team2026alayaworldv11,
  title   = {{AlayaWorld}: Interactive Long-Horizon World Modeling -- Full Technical Report (v1.1)},
  author  = {{AlayaWorld Team} and Zhang, Kaipeng and Li, Chuanhao and Zhan, Yifan and Ge, Yongtao and Yin, Yuanyang and Tan, Jiaming and He, Kang and Fan, Liaoyuan and Zhai, Mingliang and Liu, Ruicong and Xu, Xiaojie and Chu, Xuangeng and Li, Zhen and Lin, Zhengyuan and Wang, Zhixiang and Meng, Zian and Gao, Zihui},
  journal = {arXiv preprint arXiv:2608.13492},
  year    = {2026}
}

@article{evoke2026,
  title   = {{Alaya-EVOKE}: From Linear-Scaling Supervision to Endless World},
  author  = {Yin, Yuanyang and Wang, Gongxuan and Zhan, Yifan and Li, Chuanhao and Zhang, Kaipeng and Zhao, Feng},
  journal = {arXiv preprint arXiv:2608.13546},
  year    = {2026}
}

@inproceedings{wallingford2024image360,
  title     = {From an Image to a Scene: Learning to Imagine the World from a Million 360$^\circ$ Videos},
  author    = {Wallingford, Matthew and Bhattad, Anand and Kusupati, Aditya and Ramanujan, Vivek and Deitke, Matt and Kakade, Sham and Kembhavi, Aniruddha and Mottaghi, Roozbeh and Ma, Wei-Chiu and Farhadi, Ali},
  booktitle = NeurIPS,
  volume    = {37},
  year      = {2024}
}

@inproceedings{zhang2026panflow,
  title     = {{PanFlow}: Decoupled Motion Control for Panoramic Video Generation},
  author    = {Zhang, Cheng and Liang, Hanwen and Chen, Donny Y. and Wu, Qianyi and Plataniotis, Konstantinos N. and Gambardella, Camilo Cruz and Cai, Jianfei},
  booktitle = {Proceedings of the AAAI Conference on Artificial Intelligence},
  volume    = {40},
  pages     = {12385--12393},
  year      = {2026}
}

@article{ou2026holo360d,
  title   = {{Holo360D}: A Large-Scale Real-World Dataset with Continuous Trajectories for Advancing Panoramic 3D Reconstruction and Beyond},
  author  = {Ou, Jing and Cao, Zidong and Ren, Yinrui and Li, Zhuoxiao and Zhu, Jinjing and Hua, Tongyan and Zhang, Shuai and Xiong, Hui and Zhao, Wufan},
  journal = {arXiv preprint arXiv:2604.22482},
  year    = {2026}
}

@article{tan2024imagine360,
  title   = {{Imagine360}: Immersive 360 Video Generation from Perspective Anchor},
  author  = {Tan, Jing and Yang, Shuai and Wu, Tong and He, Jingwen and Guo, Yuwei and Liu, Ziwei and Lin, Dahua},
  journal = {arXiv preprint arXiv:2412.03552},
  year    = {2024}
}

@article{fang2025viewpoint,
  title   = {{ViewPoint}: Panoramic Video Generation with Pretrained Diffusion Models},
  author  = {Fang, Zixun and Zhu, Kai and Liu, Zhiheng and Liu, Yu and Zhai, Wei and Cao, Yang and Zha, Zheng-Jun},
  journal = {arXiv preprint arXiv:2506.23513},
  year    = {2025}
}

@inproceedings{liu2025dynamicscaler,
  title     = {{DynamicScaler}: Seamless and Scalable Video Generation for Panoramic Scenes},
  author    = {Liu, Jinxiu and Lin, Shaoheng and Li, Yinxiao and Yang, Ming-Hsuan},
  booktitle = CVPR,
  pages     = {6144--6153},
  year      = {2025}
}

@article{li2026cubecomposer,
  title   = {{CubeComposer}: Spatio-Temporal Autoregressive 4K 360$^\circ$ Video Generation from Perspective Video},
  author  = {Li, Lingen and Wang, Guangzhi and Li, Xiaoyu and Zhang, Zhaoyang and Dou, Qi and Gu, Jinwei and Xue, Tianfan and Shan, Ying},
  journal = {arXiv preprint arXiv:2603.04291},
  year    = {2026}
}

@inproceedings{gui2025imageworld,
  title     = {Image as a World: Generating Interactive World from Single Image via Panoramic Video Generation},
  author    = {Gui, Dongnan and Guo, Xun and Zhou, Wengang and Lu, Yan},
  booktitle = NeurIPS,
  volume    = {38},
  year      = {2025},
  doi       = {10.52202/085713-5744}
}

@article{chen2026pantheon360,
  title   = {{Pantheon360}: Taming Digital Twin Generation via 3D-Aware 360$^\circ$ Video Diffusion},
  author  = {Chen, Ting-Hsuan and Chen, Ying-Huan and Tu, Tao and Lee, Jie-Ying and Wu, Cho-Ying and Lin, Fangzhou and Zhang, Hengyuan and Paz, David and Huang, Xinyu and Guo, Yuliang and Liu, Yu-Lun and Wang, Yue and Ren, Liu},
  journal = {arXiv preprint arXiv:2605.25449},
  year    = {2026}
}

@article{li2026panoworldreal,
  title   = {{PanoWorld}: Real-World Panoramic Generation},
  author  = {Li, Haoyuan and Zhang, Dizhe and Zhou, Yuemei and Zhang, Xiangkai and Feng, Haoran and Lin, Xiaofan and Jiang, Wenjie and Du, Bo and Yang, Ming-Hsuan and Qi, Lu},
  journal = {arXiv preprint arXiv:2607.09661},
  year    = {2026}
}

@article{zhang2025worldprompter,
  title   = {{WorldPrompter}: Traversable Text-to-Scene Generation},
  author  = {Zhang, Zhaoyang and Hold-Geoffroy, Yannick and Ha{\v{s}}an, Milo{\v{s}} and Chen, Ziwen and Luan, Fujun and Dorsey, Julie and Hu, Yiwei},
  journal = {arXiv preprint arXiv:2504.02045},
  year    = {2025}
}

@article{li2026pano2world,
  title   = {{Pano2World}: End-to-End 3D Generation via Unified Multi-View Sequences},
  author  = {Li, Zhenjia and Jia, Jinrang and Shi, Yifeng},
  journal = {arXiv preprint arXiv:2607.00832},
  year    = {2026}
}

@article{su2026geniesimpanoworld,
  title   = {{Genie Sim PanoWorld}: An Infinite Indoor 3D World Generation Pipeline via Panoramic Scene Modeling and Simulation},
  author  = {Su, Yongxin and Hou, Linjie and Wang, Feng and Tang, Jialin and Li, Zhijun and Wang, Qian and Yao, Maoqing},
  journal = {arXiv preprint arXiv:2607.26646},
  year    = {2026}
}

@article{fang2026spatialcrafter,
  title   = {{SpatialCrafter}: Single Image World Modeling with Generative 3D Proxies},
  author  = {Fang, Chuan and Qiu, Lingteng and Liang, Yixun and Chen, Rui and Luo, Kunming and Zheng, Zhaohua and Bai, Tongyuan and Tian, Feipeng and Dong, Zilong and Zhou, Zihan and Tan, Ping},
  journal = {arXiv preprint arXiv:2608.27073},
  year    = {2026}
}

@article{lin2026alayarendererflash,
  title   = {Generative World Renderer at the Speed of Play},
  author  = {Lin, Guixu and Huang, Zheng-Hui and Yang, Siqi and Yang, Ming-Hsuan and Zhang, Kaipeng and Wang, Zhixiang},
  journal = {arXiv preprint arXiv:2607.18703},
  year    = {2026}
}

@article{hirschorn2026spherope,
  title   = {{SpheRoPE}: Zero-Shot Optimization-Free 360 Panorama Generation with Spherical {RoPE}},
  author  = {Hirschorn, Or and Olender, Aaron and Alshan, Eli and Ideses, Ianir and Fritz, Lior and Benaim, Sagie},
  journal = {arXiv preprint arXiv:2606.32033},
  year    = {2026}
}

@article{hong2025relic,
  title   = {{RELIC}: Interactive Video World Model with Long-Horizon Memory},
  author  = {Hong, Yicong and Mei, Yiqun and Ge, Chongjian and Xu, Yiran and Zhou, Yang and Bi, Sai and Hold-Geoffroy, Yannick and Roberts, Mike and Fisher, Matthew and Shechtman, Eli and Sunkavalli, Kalyan and Liu, Feng and Li, Zhengqi and Tan, Hao},
  journal = {arXiv preprint arXiv:2512.04040},
  year    = {2025}
}

@article{park2025spherediff,
  title   = {{SphereDiff}: Tuning-Free Omnidirectional Panoramic Image and Video Generation via Spherical Latent Representation},
  author  = {Park, Minho and Kang, Taewoong and Yun, Jooyeol and Hwang, Sungwon and Choo, Jaegul},
  journal = {arXiv preprint arXiv:2504.14396},
  year    = {2025}
}

@article{xie2025videopanda,
  title   = {{VideoPanda}: Video Panoramic Diffusion with Multi-View Attention},
  author  = {Xie, Kevin and Sabour, Amirmojtaba and Huang, Jiahui and Paschalidou, Despoina and Klar, Greg and Iqbal, Umar and Fidler, Sanja and Zeng, Xiaohui},
  journal = {arXiv preprint arXiv:2504.11389},
  year    = {2025}
}

@article{xu2025vots,
  title   = {{360VOTS}: Visual Object Tracking and Segmentation in Omnidirectional Videos},
  author  = {Xu, Yinzhe and Huang, Huajian and Chen, Yingshu and Yeung, Sai-Kit},
  journal = {IEEE Transactions on Pattern Analysis and Machine Intelligence},
  year    = {2025},
  eprint  = {2404.13953},
  archivePrefix = {arXiv}
}

@inproceedings{xu2026ucm,
  title     = {{UCM}: Unifying Camera Control and Memory with Time-Aware Positional Encoding Warping for World Models},
  author    = {Xu, Tian-Xing and Wang, Zi-Xuan and Wang, Guangyuan and Hu, Li and Zhang, Zhongyi and Zhang, Peng and Zhang, Bang and Zhang, Song-Hai},
  booktitle = {ACM SIGGRAPH 2026 Conference Papers},
  year      = {2026},
  doi       = {10.1145/3799902.3811088},
  eprint    = {2602.22960},
  archivePrefix = {arXiv}
}

@inproceedings{yan2024panovos,
  title     = {{PanoVOS}: Bridging Non-Panoramic and Panoramic Views with Transformer for Video Segmentation},
  author    = {Yan, Shilin and Xu, Xiaohao and Zhang, Renrui and Hong, Lingyi and Chen, Wenchao and Zhang, Wenqiang and Zhang, Wei},
  booktitle = {Proceedings of the European Conference on Computer Vision},
  pages     = {346--365},
  year      = {2024},
  eprint    = {2309.12303},
  archivePrefix = {arXiv}
}

@article{unterthiner2018towards,
  title   = {Towards Accurate Generative Models of Video: A New Metric \& Challenges},
  author  = {Unterthiner, Thomas and van Steenkiste, Sjoerd and Kurach, Karol and Marinier, Raphael and Michalski, Marcin and Gelly, Sylvain},
  journal = {arXiv preprint arXiv:1812.01717},
  year    = {2018},
  url     = {https://arxiv.org/abs/1812.01717}
}

@article{wang2004image,
  title   = {Image Quality Assessment: From Error Visibility to Structural Similarity},
  author  = {Wang, Zhou and Bovik, Alan C. and Sheikh, Hamid R. and Simoncelli, Eero P.},
  journal = {IEEE Transactions on Image Processing},
  volume  = {13},
  number  = {4},
  pages   = {600--612},
  year    = {2004},
  doi     = {10.1109/TIP.2003.819861}
}

@inproceedings{zhang2018unreasonable,
  title     = {The Unreasonable Effectiveness of Deep Features as a Perceptual Metric},
  author    = {Zhang, Richard and Isola, Phillip and Efros, Alexei A. and Shechtman, Eli and Wang, Oliver},
  booktitle = {Proceedings of the IEEE Conference on Computer Vision and Pattern Recognition},
  pages     = {586--595},
  year      = {2018},
  url       = {https://richzhang.github.io/PerceptualSimilarity/}
}

@article{huang2024vbenchpp,
  title   = {{VBench++}: Comprehensive and Versatile Benchmark Suite for Video Generative Models},
  author  = {Huang, Ziqi and Zhang, Fan and Xu, Xiaojie and He, Yinan and Yu, Jiashuo and Dong, Ziyue and Ma, Qianli and Chanpaisit, Nattapol and Si, Chenyang and Jiang, Yuming and Wang, Yaohui and Chen, Xinyuan and Chen, Ying-Cong and Wang, Limin and Lin, Dahua and Qiao, Yu and Liu, Ziwei},
  journal = {IEEE Transactions on Pattern Analysis and Machine Intelligence},
  year    = {2025},
  doi     = {10.1109/TPAMI.2025.3633890}
}

\end{document}